\documentclass[12pt]{article}

\usepackage{amsmath,amsfonts,bm}

\def\eqref#1{equation~\ref{#1}}

\def\1{\bm{1}}

\def\vmu{{\bm{\mu}}}
\def\vtheta{{\bm{\theta}}}

\def\va{{\bm{a}}}

\def\vm{{\bm{m}}}

\def\vu{{\bm{u}}}

\def\vw{{\bm{w}}}
\def\vx{{\bm{x}}}

\def\mI{{\bm{I}}}

\def\mV{{\bm{V}}}
\def\mW{{\bm{W}}}

\def\mPhi{{\bm{\Phi}}}

\DeclareMathAlphabet{\mathsfit}{\encodingdefault}{\sfdefault}{m}{sl}
\SetMathAlphabet{\mathsfit}{bold}{\encodingdefault}{\sfdefault}{bx}{n}

\newcommand{\Var}{\mathrm{Var}}

\usepackage{hyperref}
\usepackage{url}
\usepackage{amsthm, bm}
\usepackage{xspace}
\usepackage{amssymb}
\usepackage{multicol, multirow, booktabs, makecell}
\usepackage{caption, subcaption, graphicx}
\usepackage{dsfont}
\usepackage{amsmath}
\usepackage{times}
\usepackage{graphicx}
\usepackage{color}
\usepackage{multirow}
\usepackage[authoryear]{natbib}
\usepackage{rotating}
\usepackage{bbm}
\usepackage{latexsym}

\newcommand{\captionfonts}{\normalsize}

\makeatletter  
\long\def\@makecaption#1#2{%
  \vskip\abovecaptionskip
  \sbox\@tempboxa{{\captionfonts #1: #2}}%
  \ifdim \wd\@tempboxa >\hsize
    {\captionfonts #1: #2\par}
  \else
    \hbox to\hsize{\hfil\box\@tempboxa\hfil}%
  \fi
  \vskip\belowcaptionskip}
\makeatother   

\makeatletter
\DeclareRobustCommand\onedot{\futurelet\@let@token\@onedot}
\def\@onedot{\ifx\@let@token.\else.\null\fi\xspace}
\def\eg{\emph{e.g}\onedot} 
\def\ie{\emph{i.e}\onedot} 
\def\cf{\emph{cf}\onedot}

\makeatother

\newtheorem{theorem}{Theorem}[section]

\newtheorem{definition}{Definition}[section]

\newtheorem{property}{Property}[section]

\begin{document}
\hspace{13.9cm}1

\begin{center}
{\LARGE 
Backdoor Mitigation by Correcting the \\
Distribution of Neural Activations 
}

{ \large Xi Li$^{\displaystyle 1, \displaystyle 2}$, 
 Zhen Xiang$^{\displaystyle 1}$,
 David J. Miller$^{\displaystyle 1, \displaystyle 2}$, 
 George Kesidis$^{\displaystyle 1, \displaystyle 2}$}\\
\{xzl45,zux49,djm25,gik2\}@psu.edu\\
{
$^{\displaystyle 1}$ 
EECS,
The Pennsylvania State University, University Park, 16802, PA, USA
}\\
{
$^{\displaystyle 2}$ Anomalee Inc., State College, 16803, PA, USA
}
\end{center}
%


%
%
\begin{center} {\bf Abstract} \end{center}
Backdoor (Trojan) attacks are an important type of adversarial exploit against deep neural networks (DNNs), wherein a test instance is (mis)classified to the attacker's target class whenever the attacker's backdoor trigger is present. In this paper, we reveal and analyze an important property of backdoor attacks: a successful attack causes an alteration in the distribution of internal layer activations for backdoor-trigger instances,  compared to that for clean instances. Even more importantly, we find that instances with the backdoor trigger will be correctly classified to their original source classes if this distribution alteration is corrected.
Based on our observations, we propose an efficient and effective method that achieves post-training backdoor mitigation by correcting the distribution alteration using reverse-engineered triggers. Notably, our method does not change {\it any} trainable parameters of the DNN, but achieves generally better mitigation performance than existing methods that do require intensive DNN parameter tuning. It also efficiently detects test instances with the trigger, which may help to catch adversarial entities in the act of exploiting the backdoor.

\section{Introduction}
Deep neural networks (DNN) have shown impressive performance in many applications, but are vulnerable to adversarial attacks. 
Recently, backdoor (Trojan) attacks have been proposed against DNNs used for image classification \cite{BadNets,Targeted-Backdoor,WaNet,DBLP:journals/corr/abs-1909-02742,HiddenTrigger,invisible}, speech recognition \cite{Trojan}, text classification \cite{8836465}, point cloud classification \cite{ZhenICCV}, and even deep regression \cite{LKML21}. The attacked DNN will, with high probability, classify to the attacker's target class when a test instance is embedded with the attacker's backdoor trigger. Moreover, this is achieved while maintaining high accuracy on backdoor-free instances. Typically, a backdoor attack is launched by poisoning the training set of the DNN with a few instances embedded with the trigger and (mis)labeled to the target class.

Most existing works on backdoors either focus on improving the stealthiness of attacks \cite{Zhao_2022_CVPR, Wang_2022_CVPR}, their flexibility for launching \cite{Bai_2022_ECCV, Qi_2022_CVPR}, their adaptation for different learning paradigms \cite{Xie2020DBA, Yao_2019_CCS, Wang_2021_IJCAI}, or develop defenses for different practical scenarios \cite{Differential_Privacy, ABS, NC_blackbox, SentiNet, STRIP}. However, there are few works which study the basic properties of backdoor attacks. \cite{SS} first observed that triggered instances (labeled to the target class) are separable from clean target class instances in a feature space consisting of internal layer activations of the poisoned classifier. This property led to defenses that detect and remove triggered instances from the poisoned training set \cite{AC, CI}. As another example, \cite{zhang2022how} studied the differences between the parameters of clean and attacked classifiers, which inspired a stealthier attack with minimum degradation in accuracy on clean test instances.

In this paper, we investigate an interesting {\it distribution alteration} property of backdoor attacks. In short, the learned backdoor trigger causes a change in the distribution of internal activations for test instances with the trigger, compared to that for backdoor-free instances; 
and we theoretically demonstrate that \textbf{instances with the trigger are classified to their original source classes after such distribution alteration is reversed/corrected, with trainable parameters of the poisoned model \textit{untouched}}. Accordingly, we propose a method to mitigate backdoor attacks (post-training), such that classification accuracy on test instances both with and without the trigger will be close to the test set accuracy of a clean (backdoor-free) classifier. 
In particular, we correct distribution alteration by exploiting estimated triggers reverse-engineered by a post-training backdoor detector, \eg, \cite{NC, TNNLS}.  Thus, we propose a ``detection-before-mitigation'' defense strategy, where we first detect if a given model is backdoor-poisoned, and if so, mitigate the model with the target class(es) and the associated trigger(s) estimated by the post-training detector. 

It is important to distinguish methods that focus on backdoor \textbf{mitigation} from methods which focus on backdoor \textbf{detection}. 
Examples of the latter, including \cite{NC,TNNLS,B3D,TABOR,XiangMK20}, typically detect whether a given model is backdoor poisoned, and, if so, infer the target class(es) of the attack. Some detection methods (\eg, the ones proposed by \cite{NC, B3D, TNNLS})
are reverse-engineering based detectors, which also estimate the backdoor trigger(s) associated with the inferred target class(es).
However, if an attack is detected, these detection methods may not be able to tell whether an instance (input test sample) that is classified to the inferred target class contains the trigger; moreover, these methods do not infer the (true) original class for a backdoor-trigger image.
The goals of a backdoor mitigation method are: i) to reduce the number of backdoor-trigger test instances mis-classified to the target class(es); ii) to correctly classify these backdoor-trigger instances and iii) while maintaining relatively high accuracy on clean test instances.

Compared with existing mitigation approaches, which require tuning all of the DNN's (deep neural network's) parameters, our method achieves generally better performance and does so without changing {\it any} original, \textit{trainable} parameters of the DNN.
Also, some mitigation methods are applied irrespective of whether backdoor poisoning is detected.  These methods may unacceptably degrade clean test accuracy, and do so even when the DNN is clean (attack-free).
By contrast, our mitigation is performed only {\it after} a backdoor attack is detected (\ie, ``detection-before-mitigation''), and results in only modest drops in accuracy on clean test instances.
Moreover, while most mitigation approaches are designed to correctly classify backdoor-trigger instances without detecting whether these samples in fact contain triggers, our method 
not only corrects the decisions for these instances but also makes explicit inference of whether a test instance possesses a trigger. 
Our main \textbf{contributions} in this paper are twofold:\\
\begin{enumerate}
    \item We discover the property that backdoor attacks alter the distribution of neural activations, \ie, 
    for a backdoor-attacked DNN, the distribution of neural activations for
    backdoor-trigger instances originating from some class, $s$, deviates from the distribution for clean instances from class $s$.
    While such distribution alteration is not surprising, \textbf{we are the \textit{first} to prove that the amount of degradation in classification accuracy on backdoor-trigger instances monotonically depends on the divergence between the distributions for clean and backdoor-triggered samples}.
    \item We propose a post-training backdoor mitigation approach, based on our findings, which outperforms several state-of-the-art approaches for a variety of datasets and backdoor attack settings. \textbf{This is the \textit{first} work to mitigate backdoor attacks by correcting distribution alteration using reverse-engineered triggers, \textit{without modifying the trainable DNN model parameters}.}
\end{enumerate}

Our method offers the following advantages over existing backdoor mitigation approaches:
\begin{enumerate}
    \item Since our strategy mitigates backdoor attacks by aligning distributions without altering the trainable model parameters, our method is more \textbf{robust}, particularly when \textbf{clean instances are limited} to the defender,  compared to those that involve DNN parameter tuning (see Tab.~\ref{tab:mitigation_with_few_clean_images} and Sec.~\ref{subsec:mitigation_with_limited_clean_data}).
    \item A backdoor detection is indispensable before one applies backdoor mitigation (will be justified in Sec.~\ref{subsec:detection_before_mitigation} and Tab.~\ref{tab:mitigation_on_clean_model}). Our method is able to integrate with any reverse-engineering based detection technique. In other words, our method has strong \textbf{modularity}, and is flexible to be plugged into detection systems without much modification. 
    Hence, our method could provide \textbf{improved security over time} against evolving attacks.
\end{enumerate}

\begin{figure}[]
	\centering
	\begin{subfigure}[]{\textwidth}
		\centering
		\includegraphics[width=\textwidth]{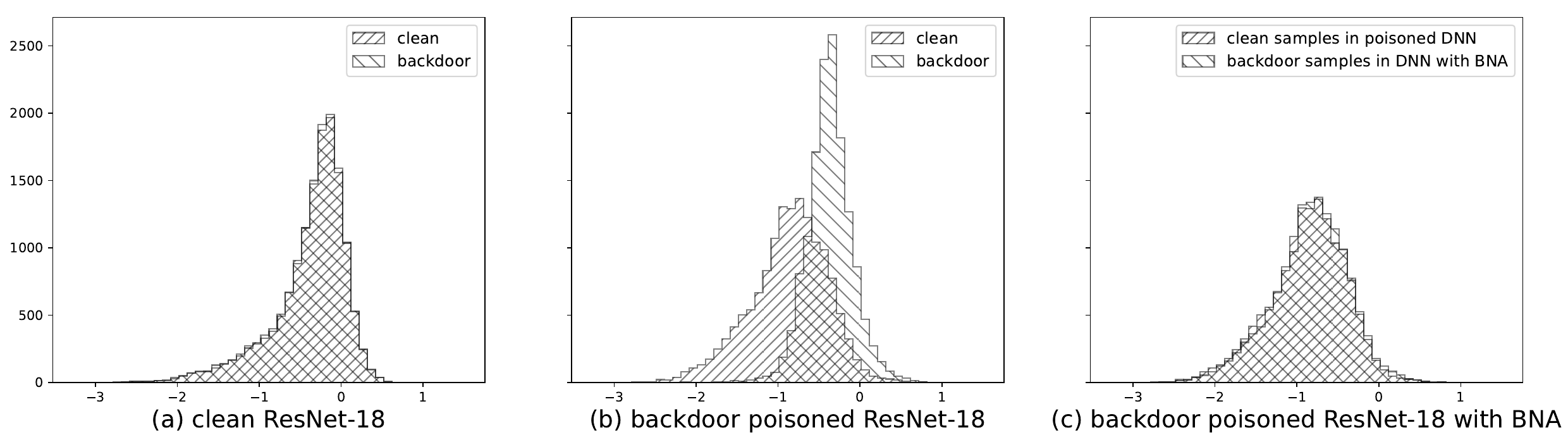}
	\end{subfigure}
	\caption{Activation distribution of a neuron in the penultimate layer of ResNet-18 trained on CIFAR-10, for instances with and without a backdoor trigger, for (a) a clean classifier and (b) a backdoor-poisoned classifier (with the same trigger). In (c), the distribution alteration in (b) is reversed by our proposed method -- most instances with the trigger will thus be correctly classified.}
	\label{fig:distributions}
\end{figure}

\section{Related Work}
There are few prior works analyzing the basic properties of backdoor attacks, \eg, the studies conducted by \cite{SS,zhang2022how}.
\cite{SS} observed that triggered instances (labeled to the target class) are separable from clean {\it target} class instances, in a feature space consisting of internal layer activations of the poisoned classifier. 
They accordingly developed a \textit{pre-training} backdoor detection system, where the detected backdoor-trigger instances are removed, and a new model is trained from scratch on the sanitized dataset. \cite{SS} thus acts like an \textit{outlier detection} system.
In contrast, we observe that backdoor attacks cause distribution alteration,
in internal layers of the DNN, between clean \textit{source} class instances and backdoor-trigger instances originating from the same (source) class.
Moreover, we demonstrate that backdoor-trigger instances are correctly classified to their classes once this distribution alteration is corrected.
We thus propose a \textit{post-training} backdoor \textit{mitigation} method based on these findings. 
In the post-training scenario, one often assumes the defender only has access to the given trained model and to a {\it small} set of clean instances, which 
generally does not include any of the training samples.  This small clean set
is inadequate for (from scratch) training an accurate, attack-free classifier. 

Existing backdoor defenses are deployed either during the DNN's training stage or post-training (but pre-deployment). 
The ultimate goal of training-stage defenses is to train an accurate, backdoor-free DNN given the possibly poisoned training set. To achieve this goal, \cite{trim_loss, huang2022backdoor, li2021anti, AC, CI, Differential_Privacy} either identify a subset of ``high-credible'' instances for training, or detect and then remove, prior to model learning, training instances that may contain a backdoor trigger. Post-training defenders, however, are assumed to have {\it no} access to the classifier's training set. Many post-training defenses aim to detect whether a given classifier has been backdoor-compromised. \cite{NC, TNNLS, DataLimited, ABS} perform anomaly detection using triggers reverse-engineered on an assumed independent clean dataset; while \cite{meta_unsup, meta_sup} train a (binary) meta classifier using ``shadow'' classifier ``exemplars'' trained with and without attack.

However, model-detection defenses are not able to mitigate backdoor attacks at test time. Thus, there is a family of post-training backdoor mitigation approaches proposed to fine-tune the classifier on the available clean dataset.
Some methods prune
neurons that may be associated with the backdoor attack \cite{Fine-Pruning, ANP, ShapPruning, CLP}; others leverage knowledge distillation to preserve the classification function only for clean instances \cite{NAD, ARGD}; and some solve a min-max optimization problem analogous to adversarial training defenses used against test-time evasion attacks \cite{hypergrad, PGD}. These defenses usually incur a significant degradation in the classifier's accuracy on clean instances, especially when the clean data available for classifier fine-tuning is insufficient. Another family of approaches are designed to detect test instances embedded with the trigger, without altering the classifier \cite{STRIP, SentiNet, Februus}. Defenses in this category may help to catch the adversarial entities in the act, but they cannot correctly classify the detected backdoor trigger instances to their original source classes. Moreover, existing methods in this category require heavy computation at test time (where rapid inferences are needed). In contrast, our mitigation framework includes both test-time trigger detection and source class inference, both with very little computation, as will be detailed in Sec. \ref{subsec:backdoor_mitigation_method}.

Neural Cleanse (NC) proposed by \cite{NC} detects backdoor attacks and then fine-tunes the classifier using instances embedded with the reverse-engineered trigger without mislabeling.
As demonstrated in Tab.~\ref{tab:distribution_divergences} in Apdx.~\ref{sec:distribution_div}, this is equivalent to distribution alignment but by altering the DNN's (trainable and non-trainable) parameters, which is \textit{not} explicitly stated in \cite{NC}. 
However, NC is \textit{not as effective as} our method in backdoor mitigation, since it tunes millions of parameters on  \textit{insufficient data} (see the last paragraph in Sec. \ref{subsec:exp_results_main} for more details). 
In contrast, we demonstrated that altering the DNN's trainable parameters is unnecessary. Instead, aligning the clean and backdoor-trigger sample distributions through straightforward transformations suffices (\cf, Thm.~\ref{thm:main}).
Moreover, NC does not detect backdoor-trigger instances during inference, unlike our method.

Most existing backdoor mitigation methods apply mitigation {\it independently} of detection, \eg, \cite{hypergrad, NAD, ARGD, ANP, MCR}. That is, they apply a mitigation method on a given model without knowing whether it is backdoor poisoned, with the expectation that the mitigation method should work well regardless of the target class(es) and associated backdoor triggers(s). 
However, mitigation may harm the model's accuracy on clean instances, especially 
when mitigation is based only on a limited amount of clean labelled data, which is common in practice, see Tab.~\ref{tab:mitigation_with_few_clean_images}.
Moreover, mitigation may waste significant computation if the given model is attack-free.
Hence, we argue that mitigation should be conducted within a ``detection-before-mitigation'' framework. In other words, one should perform backdoor mitigation only if the given model is detected as backdoor-poisoned.
This avoids a significant drop in accuracy on clean instances after mitigation (in the case where the DNN is backdoor-free).
Combined with a backdoor detection method (which may be based on embedded feature activations), our proposed mitigation method is also applicable when {\it multiple} backdoor attacks are encoded in the DNN.

\section{Distribution Alteration Property of Backdoor Attacks}\label{sec:distribution_shift_property}
In this section, we first present the \textit{activation distribution alteration property} of backdoor attacks. Then for a simplified setting, we analytically show how closing the ``gap'' between the clean-instance and backdoor-trigger instance distributions improves the accuracy in classifying backdoor-trigger instances; this will guide the design of our backdoor mitigation approach in Sec. \ref{sec:backdoor_mitigation}. 

\begin{property}\label{def:distribution_shift}
	{\bf (Activation Distribution Alteration)} For a successful backdoor attack,  different test samples embedded with the backdoor trigger will induce perturbations to the activations of an internal DNN layer that are in a similar direction. Thus, there is effectively a ``shift'' in the internal layer activation distribution for backdoor-trigger instances, compared to that for backdoor-free instances.
\end{property}


This property is easily demonstrated empirically, visually. Consider a set of clean instances from CIFAR-10 \citep{cifar10} and the {\it same} set of instances but with the backdoor trigger used by \cite{BadNets} embedded in each instance. For a ResNet-18 \citep{ResNet} classifier that was successfully attacked using this trigger, there is a {\it divergence} between the distributions of the internal layer activations induced by these two sets of instances.  This is shown in Fig. 1b for a neuron in the penultimate layer as an example. In comparison, for a clean classifier (not backdoor-attacked), the divergence between the two distributions is almost negligible as shown in Fig. 1a. 
Based on these visualizations, we ask the following question: \textit{Suppose the distribution alteration is \textbf{reversed} for each neuron, \eg, by applying a transformation to the internal activations of the triggered instances, so that the transformed distribution now closely agrees with the distribution for clean (without the backdoor-trigger) instances (see Fig. 1c). Then, following this compensation, will the classifier accurately predict the true class of origin for these backdoor-trigger instances?}


Here, we investigate this problem in a simplified binary classification setting similar to the one considered by \cite{Ilyas_NIPS_2019}. For a clean training random vector $({\bf X}, Y)$ with a uniform class prior, \ie $Y\sim\mathcal{U}\{-1, +1\}$ and with ${\bf X} | Y \sim \mathcal{N}(Y\cdot\vmu, \Sigma)$, where $\vmu\in{\mathbb R}^d$ and $\Sigma=\sigma^2\mI$, consider a backdoor attack with {\it target class} `$+1$', {\it triggered instance} ${\bf X}_b \sim \mathcal{N}(\vmu_b, \Sigma_b)$ with $\vmu_b=-\vmu+\bm{\epsilon}$, and $\Sigma_b=\sigma^{2}_b\mI$. Here, class `$-1$' is automatically the {\it source class} of ${\bf X}_b$ since there are only two classes.

With backdoor poisoning, a multi-layer perceptron (MLP) classifier is trained with one hidden layer of $J$ nodes, a batch normalization (BN) layer \footnote{Here we utilize the transformations in the BN layer to reverse the distribution alteration for simplicity. Our method does not truly rely on the existence of BN layers in the trained network, as one can always insert a BN layer between any two layers of (an already trained) network.} 
\citep{BN} followed by linear activation, and two output nodes with functions $f_{-}:{\mathbb R}^d\rightarrow{\mathbb R}$ and $f_{+}:{\mathbb R}^d\rightarrow{\mathbb R}$ corresponding to classes `$-1$' and `$+1$' respectively. An instance $\vx$ will be classified to class `$-1$' if $f_-(\vx)> f_+(\vx)$; else it will be classified to `$+1$'.

\begin{definition}\label{def:eta_erroneous}
	{\bf ($\eta$-erroneous classifier)} A classifier is said to be $\eta$-erroneous if the error rate for each class is upper bounded by $\eta$.
\end{definition}

\begin{definition}\label{def:psi_successful}
	{\bf ($\psi$-successful attack)} A backdoor attack is said to be $\psi$-successful if its attack success rate (ASR), \ie, the probability for triggered instances being (mis)classified to the attacker's target class \citep{BAsurvey}, is at least $\psi$; in our case, this means that $P[f_+({\bf X}_b) > f_-({\bf X}_b)] \geq \psi$.
\end{definition}

Given the settings above, for an arbitrary input $\vx$, the activation of the $j$-th node ($j\in\{1, \cdots, J\}$) (after BN with trained parameters $\gamma_j$ and $\beta_j$), with weight vector $\vw_j$ in the hidden layer, is:
\begin{equation}\label{eq:activation}
	a_j(\vx) = \frac{\vw_j^\top \vx - m_j}{\sqrt{v_j}} \gamma_j + \beta_j,
\end{equation}
where $m_j$ and $v_j$ respectively are the mean and variance stored by the BN layer during training on the {\it poisoned training set}. Then the activation distribution for clean source class instances 
$({\bf X}|Y=-1)\sim\mathcal{N}(-\vmu, \Sigma)$ 
is a Gaussian specified by mean ${\mathbb E}[a_j({\bf X})|Y=-1]$ and variance $\Var[a_j({\bf X})|Y=-1]$; while for triggered instances ${\bf X}_b\sim\mathcal{N}(\vmu_b, \Sigma_b)$, the activation follows a Gaussian specified by mean ${\mathbb E}[a_j({\bf X}_b)]$ and variance $\Var[a_j({\bf X}_b)]$. 
An easy way to eliminate the {\it divergence} between these two distributions is to create a classifier {\it for triggered instances} ${\bf X}_b$\footnote{These can be constructed in practice, given an estimated backdoor trigger (obtained by applying a reverse-engineering based backdoor detector, {\it e.g.}, \cite{NC,TNNLS}), by embedding the trigger in clean instances available to the defender.} 
by replacing $a_j$ in Eq. (\ref{eq:activation}) with $a^{\ast}_j(\vx)=(\vw_j^\top \vx - m^{\ast}_j)\gamma_j/\sqrt{v^{\ast}_j} + \beta_j$ for each node $j$, where (see Apdx. \ref{subsec:proof_optimal_solution} for derivation):
\begin{equation}\label{eq:mean_var_updated}
	m^{\ast}_j = \frac{\sigma_b}{\sigma} m_j + (\frac{\sigma_b}{\sigma} - 1) \vw_j^\top \vmu + \vw_j^\top \bm{\epsilon} \quad{\text{and}}\quad 
	v^{\ast}_j=\frac{\sigma_b}{\sigma} v_j.
\end{equation}
With these choices, ${\mathbb E}[a^{\ast}_j({\bf X}_b)]={\mathbb E}[a_j({\bf X})|Y=-1]$ and $\Var[a^{\ast}_j({\bf X}_b)]=\Var[a_j({\bf X})|Y=-1]$ are achieved. But here, we aim to study the quantitative relationship between the distribution divergence and the SIA metric of Def. \ref{def:sia} below. Thus, we consider an ``intermediate state'' with a classifier specified by output node functions $g_{-}(\cdot|\alpha):{\mathbb R}^d\rightarrow{\mathbb R}$ and $g_{+}(\cdot|\alpha):{\mathbb R}^d\rightarrow{\mathbb R}$, where for each output node $i\in\{-, +\}$, $g_i(\vx|\alpha)=\vu_i^\top \hat{\va}(\vx|\alpha)$ depends on a ``transition variable'' $\alpha\in[0, 1]$, with $\vu_i$ the weight vector for the original output function $f_i$. $\hat{\va}(\vx|\alpha)=[\hat{a}_1(\vx|\alpha), \cdots, \hat{a}_J(\vx|\alpha)]^\top$ is the activation vector for input $\vx$ where $\hat{a}_j(\vx|\alpha)=(\vw_j^\top \vx - \hat{m}_j(\alpha))\gamma_j/\sqrt{\hat{v}_j(\alpha)} + \beta_j$, with $\hat{m}_j(\alpha)=\alpha m_j + (1-\alpha) m^{\ast}_j$ and $\hat{v}_j(\alpha)=(\alpha \sqrt{v_j} + (1-\alpha) \sqrt{v^{\ast}_j})^2$ being the ``intermediate'' mean and variance respectively. Given these settings, our main theoretical results are presented below.

\begin{definition}\label{def:sia}
	{\bf (Source inference accuracy (SIA))} SIA is the probability that a triggered instance is classified to its original source class (\cite{InFlight}), \\
 \ie, $P[g_-({\bf X}_b|\alpha) > g_+({\bf X}_b|\alpha)]$.
\end{definition}

\begin{theorem}\label{thm:main}
	{\bf (Monotonicity of SIA with Divergence)} If the binary classifier with $f_-$ and $f_+$ is $\eta$-erroneous with $\eta<1/2$, the attack is $\psi$-successful with $\psi>1/2$, and $\sigma_b\leq\sigma$, then SIA of the modified classifier, \ie, $P[g_-({\bf X}_b|\alpha)>g_+({\bf X}_b|\alpha)]$, monotonically decreases as $\alpha\in[0, 1]$ increases.
\end{theorem}

The proof of the theorem is given in Apdx. \ref{subsec:proof_main}. Note that the assumptions for Thm. \ref{thm:main} are very mild and reasonable. For example, $\eta<1/2$ is a minimum requirement for the classifier and $\psi>1/2$ is a minimum requirement for a successful backdoor attack. Moreover, $\sigma_b\leq\sigma$ generally holds empirically since trigger embedding (\eg, consider a patch attack) typically reduces the variance of source class instances (while additive attacks do not change the variance).
Also note that $\alpha$ merely gives a way of quantifying distribution 
divergence for purpose of analysis.
According to these results, the core part of our proposed backdoor mitigation approach should be to find a modified classifier $g(\cdot|{\bm \Theta})$ by minimizing (\eg, using sub-gradient methods) a measure of distribution divergence over a well-chosen {\it set} of parameters, ${\bm \Theta}$. This approach is next explicated. 

\begin{figure}
	\centering
	\begin{subfigure}[]{\textwidth}
		\centering
		\includegraphics[width=.96\textwidth]{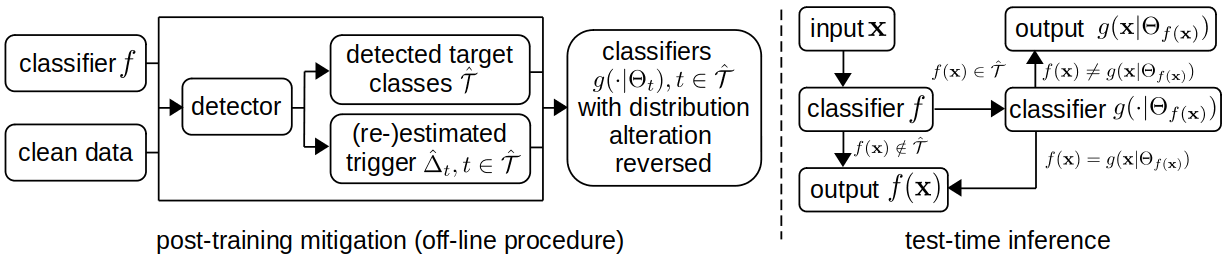}
	\end{subfigure}
	\caption{Illustration of our backdoor mitigation framework with a test-time inference rule.}
	\label{fig:procedure}
\end{figure}

\section{Reversing Distribution Alteration for Backdoor Mitigation}\label{sec:backdoor_mitigation}
\subsection{Problem Description}\label{subsec:backdoor_mitigation_problem}
{\bf Threat model.} For input space ${\mathcal X}$ and label space ${\mathcal C}$, a classifier that has been successfully backdoor-attacked will predict to the attacker's target class $t^{\ast}\in{\mathcal C}$ when a test instance $\vx\in{\mathcal X}$ is embedded with the backdoor trigger using an incorporation function $\Delta:{\mathcal X}\rightarrow{\mathcal X}$. In addition to this ``{\it all-to-one}'' setting, we also consider the ``{\it all-to-all}'' setting where a test instance from any class $c\in{\mathcal C}$ will be (mis)classified to class $(c+1)mod|{\mathcal C}|$ when it is embedded with the trigger \citep{BadNets}.

{\bf Defender's goals.} Given a trained classifier $f:{\mathcal X}\rightarrow{\mathcal C}$ that may possibly be attacked, the defender aims to mitigate possible attacks by producing a mapping $\hat{f}:{\mathcal X}\rightarrow{\mathcal C}$ which (a) has high accuracy in classifying clean instances; (b) when there is a backdoor attack, classifies triggered instances to their original source class, as though there is no trigger embedded, \ie, achieves a high SIA; and c) {\it detects} whether or not a test sample contains a backdoor trigger.

{\bf Defender's assumptions.} We consider a {\it post-training} scenario where the defender has {\it no access} to the training set of the classifier. The defender does possess an independent clean dataset, but this dataset is {\it too small} to train an accurate classifier from scratch, and even too small to effectively fine-tune the full set of classifier parameters \citep{Fine-Pruning, hypergrad, NC}. The defender has full (white box) access to the classifier, but does not know whether it has been attacked and, if so, does not know the trigger pattern that was used, \ie, the defense is unsupervised -- we will leverage existing post-training detectors to determine if the classifier was attacked and, if so, to estimate the target class(es) of the attack and the backdoor trigger.

{\bf The ``detection-before-mitigation'' scenario}: We propose that backdoor detection should generally be performed before mitigation. That is, one should first apply a backdoor detection method on the given model and perform backdoor mitigation on it {\it only} if it is detected as backdoor poisoned. 
Otherwise, backdoor mitigation may harm the accuracy of the model and is a waste of computation if there is no attack.
On the other hand, if there is an attack, utilizing the detection results (\eg, the detected target class(es)) helps to reduce the degradation in the classifier's accuracy on clean test data brought about by mitigation. 
(See the experimental results of Sec.~\ref{subsec:detection_before_mitigation} and Tab.~\ref{tab:mitigation_on_clean_model}.)
Our method indeed mitigates only when a backdoor is detected, and exploits knowledge of the detected target class(es), as well as the estimated backdoor trigger, produced
by post-training detectors such as \cite{NC,TNNLS}.

\subsection{Method}\label{subsec:backdoor_mitigation_method}

\textbf{Key idea}: The principle behind our mitigation method is simple: A backdoor-trigger instance will be correctly classified to its original source class by the poisoned model if the model is altered in such a way as to follow the same distribution as the clean source class instances in each internal layer feature space of the model (as shown in Fig.~\ref{fig:distributions} and proved by Thm. \ref{thm:main}). 
For this purpose there is \textit{no need to modify the trainable parameters}. 
We align distributions through simple transformations, \eg, those used in batch normalization, on internal layer feature maps, produced for clean samples that are embedded with the estimated backdoor trigger (heretofore referred to as ``backdoor-trigger instances''). The transformation parameters are optimized by minimizing the divergence between the distributions of clean instances and triggered instances.
To approximate the distributions, we calculate the histogram of (transformed) feature maps in each internal layer.
Our mitigation framework, along with the test-time detection rule, is visually summarized in Fig. \ref{fig:procedure}. A detailed explanation will follow the introduction of our mitigation strategy.

Now we elaborate our mitigation strategy.
Based on Thm. \ref{thm:main}, it would seem that a good mitigation approach involves modifying the classifier $f$,
\ie, creating a new classifier $g(\cdot|{\bm \Theta}):{\mathcal X}\rightarrow{\mathcal C}$ from $f$ by applying a transformation function $h_{j,l}(\cdot|\vtheta_{j,l}):{\mathbb R}\rightarrow{\mathbb R}$ to the activation of each neuron $j\in\{1, \cdots, J_l\}$ in each layer $l\in\{1, \cdots, L\}$. The transformation parameters ${\bm \Theta}=\{\vtheta_{j,l}\}$ should be jointly chosen so as to minimize the aggregation (\eg, sum) of the divergences between the distributions $q_{j,l}(\vtheta_{<l} \cup \vtheta_{j,l})$ obtained using $h_{j,l}(\hat{z}_{j,l}(\Delta({\bf X})|\vtheta_{<l})|\vtheta_{j,l})$ (\ie, distributions of the transformed activations of backdoor-trigger samples) and the target distributions $p_{j,l}$ for $z_{j,l}({\bf X})$ (\ie, distributions of the activations of clean samples) for $\forall j,l$, where ${\bf X}$ follows the clean data distribution, \ie:
\begin{align}\label{eq:opt_raw}
	\underset{{\bm \Theta}=\{\vtheta_{j,l}\}}{\text{minimize}} \quad \sum_{j, l} D_k \big(p_{j,l} || q_{j,l}(\vtheta_{<l} \cup \vtheta_{j,l}) \big)
\end{align}
where: $z_{j,l}:{\mathcal X}\rightarrow{\mathbb R}$ and $\hat{z}_{j,l}:{\mathcal X}\rightarrow{\mathbb R}$ are activation functions for neuron $j$ in layer $l$ for classifiers $f$ and $g(\cdot|{\bm \Theta})$ respectively; $\vtheta_{<l}=\{\vtheta_{j,l'}|l'<l\}$ represents all transformation parameters prior to layer $l$; $D_k(p||q):={\mathbb E}_q[k(p/q)]$ for a convex function $k:[0, \infty)\rightarrow{\mathbb R}$ satisfying $k(1)=0$ and belonging to the family of $f$-divergences for any distributions $p$ and $q$ \cite{Ali1966AGC}.

However, in practice, we have the following challenges.
\textbf{Challenge 1}: Unlike the distributions of clean samples $\{p_{j,l}\}$, which can be simply approximated by feeding the small number of clean samples possessed by the defender to the poisoned model\footnote{As previously discussed, mitigation methods should only be applied to a model if it has been detected as poisoned.} and calculating the histograms of internal activations, the distributions of backdoor-trigger samples $\{q_{j,l}\}$ are unknown. 
\textbf{Challenge 2}: The density form for the activation of backdoor-trigger samples $z_{j,l}(\Delta({\bf X}))$ may get altered by the trigger $\Delta$ and will likely be different from the density form for the activation of clean samples $z_{j,l}({\bf X})$; moreover,  both will likely be non-Gaussian. Thus, minimizing Eq.~(\ref{eq:opt_raw}) is not trivial. One cannot align the distributions by \eg,  simply matching the mean and variance.


To address Challenge 1, we approximate the distributions of true backdoor-triggers samples by those of defender's samples that are embedded with the trigger(s) estimated by a post-training detector. 
This can be accomplished with widely used post-training reverse-engineering based backdoor detection (RED) approaches which have the same assumption as in Sec.~\ref{subsec:backdoor_mitigation_problem}, \eg, the ones proposed by \cite{NC, DeepInspect, ABS, TNNLS}. 
These REDs investigate whether the classifier $f$ is compromised by a backdoor attack and if so, infer the source and target classes and estimate the associated backdoor trigger(s)\footnote{Note that these REDs can be the backdoor detectors used prior to applying mitigation methods. Thus, there is no additional computation cost involved.}.  

To solve a broad range of attack settings, \eg, all-to-one and all-to-all attacks, we apply the detection methods in a general way following \cite{TNNLS}.
We first reverse-engineer a trigger by solving an optimization problem defined on the clean set to get a detection statistic for each ordered {\it putative class pair} $(s, t)\in{\mathcal C}\times{\mathcal C}$. A statistic which \cite{TNNLS} suggests is (the reciprocal of) the estimated perturbation size inducing high (mis)classifications from $s$ to $t$. For \cite{NC}, it is the estimated patch size inducing high (mis)classifications from $s$ to $t$. Then we apply the anomaly detection approach in \cite{NC}, based on the MAD criterion \cite{MAD}, to all the obtained statistics to find {\it all} the {\it outlier} statistics. We denote the set of detected class pairs associated with these outlier statistics as $\hat{\mathcal P}$, and denote $\hat{\mathcal T}=\{t\in{\mathcal C}~|~\exists s\in{\mathcal C}~ {\rm s.t.}~ (s, t)\in \hat{\mathcal P}\}$ as the set of detected target classes. 

For {\it each} $t\in\hat{\mathcal T}$, we (re-)estimate a trigger
$\hat{\Delta}_t$ (as a {\it surrogate} for the true backdoor trigger, which is unknown) using clean instances from {\it all} detected source classes\footnote{More reliable trigger estimation can be achieved in this way for a detected target class, compared with estimating the trigger based on only one (source, target) class pair.} $\hat{\mathcal S}(t)=\{s\in{\mathcal C}|(s, t)\in\hat{\mathcal P}\}$. Then, {\it for each detected target class} $t\in\hat{\mathcal T}$, we construct a classifier $g(\cdot|{\bm \Theta}_t)$ by solving the distribution divergence minimization problem using the (re-)estimated $\hat{\Delta}_t$.

Now we address Challenge 2,
which is \textit{critical} to the estimation of ${\bm \Theta}_t$ using the reverse-engineered trigger $\hat{\Delta}_t$ for each detected target class $t\in\hat{\mathcal T}$. For simplicity, we will drop the subscript $t$ below without loss of generality. Our main goals are: (a) specifying the structure of the transformation function $h_{j,l}$ with its associated parameters ${\vtheta_{j,l}}$, (b) empirical estimation of the distribution divergence in Eq. (\ref{eq:opt_raw}) using a clean dataset (\ie, the subset of clean instances from classes in $\hat{\mathcal S}(t)$ for each detected class $t$), and (c) choosing the convex function $k$ to specify the divergence form. For (a), we consider the following transformation function with parameters ${\vtheta_{j,l}}=\{\mu_{j,l}, \sigma_{j,l}, \upsilon_{j,l}, \omega_{j,l}\}$:
\begin{equation}\label{eq:transform_wo_bn}
	h_{j,l}(z) = \max \{ \min \{ \frac{z-\mu_{j,l}}{\sigma_{j,l}}, \omega_{j,l}\}, \upsilon_{j,l}\}
\end{equation}
where $\mu_{j,l}$ and $\sigma_{j,l}$ specify the location and scale of the activation distribution, respectively, while $\upsilon_{j,l}, \omega_{j,l}$ control the shape of the tail of the distribution. For goal (b), we quantize the real line into $M$ intervals ${\mathcal I}_1=(-\infty, b_1), {\mathcal I}_2=[b_1, b_2), \cdots, {\mathcal I}_M=[b_{M-1}, \infty)$, for $M$ sufficiently large. Then the distribution divergence in Eq. (\ref{eq:opt_raw}) for each node $j$ and layer $l$ is computed on discrete distributions $\hat{p}_{j,l}$ and $\hat{q}_{j,l}$ over these intervals. Specifically, the discrete distributions are estimated using a subset ${\mathcal D}_{t}$ of instances from classes $\hat{\mathcal S}(t)$, with the probabilities for interval ${\mathcal I}_i$ computed by:
\begin{equation}\label{eq:discrete_dists}
\begin{split}
    \hat{p}_{j,l}^{(i)} &= \frac{1}{|{\mathcal D}_t|} \sum_{\vx\in{\mathcal D}_t} {\mathds 1} [z_{j,l}(\vx) \in {\mathcal I}_i] \,\,\,\,{\text{and}} \\
    \hat{q}_{j,l}^{(i)} &= \frac{1}{|{\mathcal D}_t|} \sum_{\vx\in{\mathcal D}_t} {\mathds 1} [h_{j,l}(\hat{z}_{j,l}(\hat{\Delta}_t({\bf X})|\vtheta_{<l})|\vtheta_{j,l}) \in {\mathcal I}_i].
\end{split}
\end{equation}

To ensure that the distribution divergence is differentiable with reference to the parameters, such that it can be minimized using (\eg) gradient descent, we approximate the non-differentiable indicator function ${\mathds 1}[\cdot]$ in Eq. (\ref{eq:discrete_dists}) using differentiable functions such as the sigmoid, \ie, we redefine:
\begin{equation}\label{eq:approx}
	{\mathds 1}[z\in {\mathcal I}_i] = sigmoid(\tau(z-b_{i-1})) - sigmoid(\tau(z-b_i))
\end{equation}
where $\tau$ is a scale factor controlling the error of approximation. For ${\mathcal I}_1$ and ${\mathcal I}_M$, which have semi-infinite support, we use a single sigmoid in Eq. (\ref{eq:approx}). The choice of the intervals and $\tau$ is not critical to the performance, as long as the length of the finite intervals is sufficiently small, as will be shown in Tab.~\ref{tab:t&bin_size} in Sec.~\ref{sec:exp}. Finally, for goal (c), we consider several different divergence forms including the total variation (TV) divergence with $k(r)=|r-1|/2$, the Jensen-Shannon (JS) divergence with $k(r)=r\log\frac{2r}{r+1}+\log\frac{2}{r+1}$, and the Kullback-Leibler (KL) divergence with $k(r)=r\log r$. The choice of the divergence form is also not critical to the mitigation performance (see Apdx.~\ref{sec:choice_loss_func}).

We now provide a detailed explanation of our backdoor mitigation framework, which is visually summarized in Fig.~\ref{fig:procedure}.
For any test input $\vx\in{\mathcal X}$, if classifier $f$ is deemed attack-free, \ie, $\hat{\mathcal P}=\emptyset$, the classification output under our mitigation framework will be $\hat{f}(\vx)=f(\vx)$. 
Otherwise, if $f(\vx)\in{\mathcal C}\setminus{\hat{\mathcal T}}$, we trust the class decision and set $\hat{f}(\vx)=f(\vx)$ both because $\vx$ is unlikely to possess a trigger and because a successful attack should not degrade the classifier's accuracy on clean instances. 
However, if $f(\vx)=t\in\hat{\mathcal T}$, there are two main possibilities: 1) $\vx$ is a clean instance truly from class $t$; 2) $\vx$ is classified to class $t$ due to the presence of the trigger. To distinguish these two cases, we feed $\vx$ to the optimized $g(\cdot|{\bm \Theta}_t)$. 
If $g(\vx|{\bm \Theta}_t)\neq f(\vx)$, $\vx$ likely contains a trigger, and thus we should set $\hat{f}(\vx)=g(\vx|{\bm \Theta}_t)$, which is likely the original source class of $\vx$ based on our theoretical results. 
Note that in the test-time inference procedure above, the major (additional) computation for both backdoor trigger instance detection and source class inference is a forward propagation, feeding $\vx$ to $g(\cdot|{\bm \Theta}_t)$, which is comparable to the computation required for classification using $f$. 
Moreover, such additional computation occurs only if an attack is detected and $f(\vx)=t$; thus, our test-time inference is very efficient.

\section{Experiments}\label{sec:exp}

\subsection{Experiment Setup} \label{subsec:exp_setup}
\textbf{Datasets}: Our main experiments are conducted on the benchmark CIFAR-10 dataset, which contains 60,000 $32\times32$ color images from 10 classes, with 5,000 images per class for training and 1,000 images per class for testing (\cite{cifar10}). 
We also show the effectiveness of our proposed mitigation framework on other benchmark datasets including GTSRB (\cite{GTSRB}), CIFAR-100 (\cite{cifar10}), ImageNette (\cite{imagewang}), TinyImageNet \cite{TinyImageNet}, and VGGFace2 \cite{vggface2}. 
Details of these datasets can be found in Apdx.~\ref{sec:datasets}.
Data allocation in our experiments strictly follows the assumptions in Sec. \ref{subsec:backdoor_mitigation_problem}. For each dataset, we randomly sample 10\% of the test set to form the small, clean dataset $\mathcal{D}_{\text{Defense}}$ assumed for the defender. The remaining test instances, denoted by $\mathcal{D}_{\text{Test}}$, are reserved for performance evaluation.
\\
\textbf{Attack settings}:
In this paper, we consider standard backdoor attacks launched by poisoning the training set of the classifier \citep{BadNets, Targeted-Backdoor}. In particular, we consider both the \textit{all-to-one} (\textbf{A2O}) attacks and the \textit{all-to-all} (\textbf{A2A}) attacks in our main experiments on CIFAR-10. 
For A2O attacks on CIFAR-10, we arbitrarily choose class 9 as the target class; while for A2A attacks, as described in Sec. \ref{subsec:backdoor_mitigation_problem}, triggered instances from any class $c\in{\mathcal C}$ are supposed to be (mis)classified to class $(c+1)mod|{\mathcal C}|$. For each attack setting, we consider the following triggers:
1) a $3\times3$ random patch (\textbf{BadNet}) with a randomly selected location (fixed for all triggered images for each attack) used in \cite{BadNets};
2) an additive perturbation (with size 2/255) resembling a chessboard (\textbf{CB}) used in \cite{TNNLS};
3) a single pixel (\textbf{SP}) perturbed by 75/255 with a randomly selected location (fixed for all triggered images for each attack) used by \cite{SS}; 
4) invisible triggers generated with $l_0$ and $l_2$ norm constraints (\textbf{$l_0$ inv} and \textbf{$l_2$ inv} respectively) proposed by \cite{invisible};
5) a warping-based trigger (\textbf{WaNet}) proposed by \cite{WaNet};
6) a Hello Kitty blending trigger (\textbf{Blend}) used by \cite{Targeted-Backdoor}; 
7) a trigger generated by the horizontal sinusoidal function (\textbf{SIG}) defined in \cite{SIG}. 
Details for generating these triggers are deferred to Apdx.~\ref{sec:attack_settings}. We randomly created 5 attacks for each attack setting {\it e.g.} by
randomly locating the trigger. 
We also evaluated against the ``label-consistent" (\textbf{CL}) backdoor attack proposed by \cite{CL} on the CIFAR-10 dataset, which only embeds the backdoor trigger into the target class training samples. Details are given in Apdx. \ref{sec:attack_settings}.
For experiments on the other five datasets, we only consider A2O attacks for a subset of triggers where sufficiently high success rate can be achieved. For each dataset, we create one attack for each trigger being considered. A2A attacks are not considered for these datasets since there is insufficient data per class for them to achieve successful attacks. More details about the attacks, including the number of backdoor-trigger images used for poisoning and the target class selected to create A2O attacks for the five datasets other than CIFAR-10, are shown in Apdx. \ref{sec:attack_settings}. \\
\textbf{Performance evaluation metrics}: 1) The attack success rate (\textbf{ASR}) is the fraction of clean instances in $\mathcal{D}_{\text{Test}}$ (mis)classified to the designated target class when the backdoor trigger is embedded. 2) The clean test accuracy (\textbf{ACC}) is the DNN's accuracy on $\mathcal{D}_{\text{Test}}$ without trigger embedding. 3) The \textbf{SIA} (Def \ref{def:sia}) is the fraction of clean instances in $\mathcal{D}_{\text{Test}}$ classified to the original source class when the trigger is embedded. For a successful backdoor attack, ASR and ACC should be high, while SIA should be low. For a successful mitigation approach, the resulting ASR should be low, while ACC and SIA should be high.\\
\textbf{Training settings}: We train one classifier for each attack to evaluate our mitigation approach against existing ones. Training configurations, including the DNN architecture, batch size, number of epochs, etc., are detailed in Tab.~\ref{tab:training_config} in Apdx.~\ref{sec:training_settings}. Data augmentation choices, including random cropping and horizontal flipping, are applied to each training instance. As shown in Tab. \ref{tab:comparison}, the defenseless ``vanilla'' classifiers being attacked 
achieve high ACC but suffer high ASR and low SIA
(averaged over the five attacks we created) for all trigger types and for both A2O and A2A settings,
\ie, the attacks are all successful and hence adequate for performance evaluation.\\
\textbf{Hyper-parameter settings}:
We compare our mitigation approach (named `Batch Normalization Alteration' (\textbf{BNA}) in the sequel) with six well-known and/or state-of-the-art methods, including NC(\cite{NC}), NAD(\cite{NAD}), I-BAU(\cite{hypergrad}), ANP(\cite{ANP}), ARGD(\cite{ARGD}), and MCR\cite{MCR}. 
For MCR, in their original paper, the defender is assumed to have access to two poisoned models, which may be impractical. Thus, we fine-tune the given model on the defender's dataset and use it as the second model (which is also suggested in their paper).
For all these other methods, we used their officially posted code for implementation. 
For BNA, following Sec. \ref{subsec:backdoor_mitigation_method}, we first perform detection by reverse-engineering a backdoor trigger for each class pair using objective functions from \cite{NC, TNNLS} and then feed the statistics obtained based on the estimated trigger to an anomaly detector. 
Our anomaly detector is based on MAD, which is a classical approach also used by \cite{NC, DeepInspect, DataLimited}. Here, we set the detection threshold at ``7-MAD'' which easily catches all the backdoor class pairs. More details, including pattern estimation and detection statistics are shown in Apdx. \ref{sec:backdoor_detection}. 
Then, for each detected target class, we solve the divergence minimization problem to optimize the transformation functions using learning rate 0.01 for 10 epochs. 
Since our mitigation method applies simple transformations which are also used in BN, we consider model structures that contain BN layers (which is very common) for simplicity. But note that the proof of monotonicity of SIA with distribution divergence (Thm.~\ref{thm:main}) and our method (Sec.~\ref{subsec:backdoor_mitigation_method}) do not truly rely on the presence of BN layers -- one can always insert a BN layer between any two given layers of the trained network.
If a neuron is followed by a BN, instead of applying an additional transformation function $h_{j,l}$, we treat the mean and standard deviation of BN as the parameters $\mu_{j,l}$ and $\sigma_{j,l}$ associated with $h_{j,l}$ respectively. 
We optimize the mean and standard deviation by minimizing distribution divergence for all the BN layers.
In Sec.~\ref{subsec:exp_results_main}, we only show results for BNA with the total variation divergence. Results for KL-divergence and JS-divergence are deferred to Apdx.~\ref{sec:choice_loss_func}. 
To compute the divergence, we use the ``interval trick'' (Eq. (\ref{eq:discrete_dists})) to obtain the discrete empirical distribution. 
For simplicity, we let all finite intervals, ${\mathcal I}_i=[b_{i-1}, b_i), i=1,\cdots,M$, have the same length $\Delta b=0.1$.
For each neuron, we set $b_{\min}$ and $b_{\max}$ as the minimum and maximum activations, respectively, when feeding in clean instances from $\mathcal{D}_{\text{Defense}}$ to the poisoned classifier $f$. 
Then, the number of intervals is $M=\lceil \frac{b_{\max}-b_{\min}}{\Delta b} \rceil$; and all intervals can be specified by $b_0=b_{\min}$ and $b_i = b_{i-1}+\Delta b$. 
Finally, the scale factor in Eq. (\ref{eq:approx}) is set to $\tau=150$, which is obtained by line search to minimize the total variation between the ``soft'' distribution and the empirical one on $\mathcal{D}_{\text{Defense}}$. 
In fact, the choices for $\Delta b$ and $\tau$ (over reasonable ranges) have little impact on our mitigation performance, as shown in Tab. \ref{tab:t&bin_size}.\\

\begin{table}[h!]
\setlength{\tabcolsep}{4pt}
\centering
\resizebox{.99\textwidth}{!}{
\begin{tabular}{cccccccccccccc}
\toprule
\hline
\multirow{2}{*}{\makecell[c]{Trigger \\ type}} &  &  \multicolumn{2}{c}{BadNet} & \multicolumn{2}{c}{CB} & \multicolumn{2}{c}{$l_0$ inv} & \multicolumn{2}{c}{$l_2$ inv} & \multicolumn{2}{c}{SP} & \multicolumn{2}{c}{WaNet} \\
\cline{3-14}
&  & A2O & A2A & A2O & A2A & A2O & A2A & A2O & A2A & A2O & A2A & A2O & A2A \\ 
\hline
\multirow{3}{*}{\makecell[c]{Vanilla}}
& ACC & 0.9122 & 0.9121 & 0.9135 & 0.9098 & 0.9135 & 0.9131 & 0.9130 & 0.9126 & 0.9138 & 0.9060 & 0.9032 & 0.8994 \\
& ASR & 0.9573 & 0.8658 & 0.9685 & 0.8692 & 0.9989 & 0.8973 & 0.9889 & 0.8620 & 0.8912 & 0.8550 & 0.9153 & 0.8216 \\
& SIA & 0.0397 & 0.0432 & 0.0293 & 0.0257 & 0.0010 & 0.0151 & 0.0107 & 0.0194 & 0.1016 & 0.0593 & 0.0772 & 0.0714 \\ 
\hline
\multirow{3}{*}{NC}
& ACC & 0.8797 & 0.8762 & 0.8735 & 0.8776 & 0.8835 & 0.8767 & 0.8750 & 0.8690 & 0.8854 & 0.8614 & 0.8748 & 0.8756 \\ 
& ASR & 0.0130 & \textbf{0.0154} & \textbf{0.0064} & 0.0155 & 0.0120 & 0.0150 & 0.0080 & 0.0179 & 0.0335 & 0.0188 & 0.0144 & 0.1381 \\ 
& SIA & 0.8532 & 0.8614 & 0.8312 & 0.8597 & 0.8654 & 0.8650 & 0.7932 & 0.8254 & 0.8362 & 0.8477 & 0.8231 & 0.7183 \\
\hline
\multirow{3}{*}{I-BAU}
& ACC & 0.8500 & 0.8758 & 0.8812 & 0.8719 & 0.8452 & 0.8800 & 0.8825 & 0.8726 & 0.8666 & 0.8745 & 0.8777 & 0.8700 \\ 
& ASR & \textbf{0.0094} & 0.0164 & 0.1973 & 0.0811 & 0.0091 & 0.0133 & 0.2600 & 0.3353 & 0.0172 & \textbf{0.0154} & 0.1339 & 0.1253 \\ 
& SIA & 0.8301 & 0.8583 & 0.6399 & 0.7756 & 0.8277 & 0.8673 & 0.5549 & 0.4928 & 0.8479 & 0.8609 & 0.7059 & 0.7269 \\
\hline
\multirow{3}{*}{ANP}
& ACC & 0.8644 & 0.8492 & 0.8241 & 0.8577 & 0.8455 & 0.8648 & 0.8345 & 0.8421 & 0.8195 & 0.8411 & 0.8298 & 0.8607 \\ 
& ASR & 0.0474 & 0.1199 & 0.3351 & 0.0927 & 0.0836 & 0.1326 & 0.4703 & 0.2648 & 0.1229 & 0.0495 & 0.0263 & 0.0835 \\ 
& SIA & 0.8184 & 0.7205 & 0.4587 & 0.7168 & 0.7697 & 0.7324 & 0.3351 & 0.4976 & 0.7060 & 0.7942 & 0.7368 & 0.7451 \\
\hline
\multirow{3}{*}{NAD}
& ACC & 0.8814 & 0.8819 & 0.8800 & \textbf{0.8908} & 0.8958 & \textbf{0.9047} & 0.8991 & \textbf{0.8781} & 0.8813 & 0.8761 & 0.8592 & \textbf{0.8963} \\
& ASR & 0.0193 & 0.7132 & 0.0871 & 0.0681 & 0.0356 & 0.0457 & 0.0254 & 0.0191 & 0.0667 & 0.0647 & 0.0571 & 0.1056 \\
& SIA & 0.8498 & 0.1520 & 0.7711 & 0.8084 & 0.8504 & 0.8534 & 0.8221 & 0.8337 & 0.8123 & 0.8082 & 0.7710 & 0.7773 \\
\hline
\multirow{3}{*}{ARGD}
& ACC & 0.8689 & 0.8482 & 0.8800 & 0.8774 & 0.8880 & 0.8885 & 0.8669 & 0.8583 & 0.8899 & 0.8728 & 0.8739 & 0.8755 \\ 
& ASR & 0.0368 & 0.0839 & 0.0099 & \textbf{0.0117} & 0.0079 & 0.0122 & 0.0125 & 0.0179 & 0.0955 & 0.0452 & 0.0111 & 0.0362 \\ 
& SIA & 0.8217 & 0.7544 & 0.8657 & 0.8690 & 0.8725 & 0.8786 & 0.8168 & 0.8297 & 0.7934 & 0.8295 & 0.8283 & 0.8241 \\
\hline
\multirow{3}{*}{\makecell[c]{MCR}}
& ACC & 0.8751 & 0.8840 & 0.8126 & 0.8618 & 0.8534 & 0.8834 & 0.8688 & 0.8481 & 0.8886 & 0.8613 & 0.8808 & 0.8661 \\
& ASR & 0.1447 & 0.1001 & 0.6015 & 0.1004 & 0.2900 & \textbf{0.0000} & 0.9769 & 0.0971 & 0.0136 & 0.1166 & 0.0268 & 0.0974 \\
& SIA & 0.7572 & 0.0811 & 0.3151 & 0.1163 & 0.0927 & 0.1000 & 0.0210 & 0.3524 & 0.8570 & 0.2239 & 0.7999 & 0.7789 \\
\hline
\multirow{3}{*}{\makecell[c]{BNA \\ (ours)}}
& ACC & \textbf{0.9032} & \textbf{0.8951} & \textbf{0.9072} & 0.8615 & \textbf{0.9068} & 0.8944 & \textbf{0.9005} & 0.8638 & \textbf{0.9058} & \textbf{0.8921} & \textbf{0.8945} & 0.8792 \\ 
& ASR & 0.0139 & 0.0189 & 0.0127 & 0.0202 & \textbf{0.0033} & 0.0111 & \textbf{0.0042} & \textbf{0.0168} & \textbf{0.0104} & 0.0225 & \textbf{0.0041} & \textbf{0.0191} \\ 
& SIA & \textbf{0.8835} & \textbf{0.8841} & \textbf{0.8787} & \textbf{0.8820} & \textbf{0.8924} & \textbf{0.8942} & \textbf{0.8383} & \textbf{0.8522} & \textbf{0.8863} & \textbf{0.8811} & \textbf{0.8530} & \textbf{0.8607} \\
\hline
\bottomrule
\end{tabular}
}
\caption{Average ACC, ASR, and SIA for BNA, compared with NC, NAD, I-BAU, ANP, and ARGD, against all the created attacks applied to ResNet-18 trained on the CIFAR-10 dataset.  Best performances are indicated in bold.}
\label{tab:comparison}
\end{table}

\begin{table}[h!]
\centering
\scriptsize
\begin{tabular}{cccccccccc}
\toprule
\hline
&  & Vanilla & NC & I-BAU & ANP & NAD & ARGD & MCR & BNA \\
\hline
\multirow{3}{*}{CL}
& ACC & 0.9062 & \textbf{0.9061} & 0.1735 & 0.8998 & 0.3354 & 0.2362 & 0.8829 & 0.8967 \\ 
& ASR & 0.9304 & 0.1444 & 0.4940 & 0.4632 & 0.0636 & \textbf{0.0581} & 0.7756 & 0.0594 \\ 
& SIA & 0.0667 & 0.7558 & 0.0896 & 0.4977 & 0.2985 & 0.2349 & 0.1932 & \textbf{0.8057} \\
\bottomrule
\end{tabular}
\caption{ACC, ASR, and SIA for BNA, compared with those of NC, NAD, I-BAU, ANP, and ARGD, against the ResNet-18 trained on the CIFAR-10 dataset poisoned by the label-consistent (CL) backdoor attack.}
\label{tab:mitigation_on_CL_attack}
\end{table}

\begin{table}[h!]
\centering
\resizebox{.99\textwidth}{!}{
\begin{tabular}{cccccccccccccccc}
\toprule
\hline
& \multicolumn{4}{c}{VGGFace2} & \multicolumn{8}{c}{CIFAR-10} \\
& Vanilla & NC & I-BAU & BNA & Vanilla & NC & I-BAU & ANP & NAD & ARGD & MCR & BNA \\
\hline
ACC & 0.8989 & \textbf{0.8967} & 0.6828 & 0.8917 & 0.9122 & 0.8848 & 0.8539 & 0.6463 & 0.8731 & 0.4946 & 0.8650 & \textbf{0.9026} \\
ASR & 0.9771 & 0.9693 & 0.9737 & \textbf{0.0046} & 0.9573 & 0.2531 & 0.4175 & \textbf{0.0117} & 0.8880 & 0.0227 & 0.8078 & 0.0185 \\
SIA & 0.0216 & 0.0294 & 0.0196 & \textbf{0.8889} & 0.0397 & 0.6842 & 0.5148 & 0.6324 & 0.1007 & 0.4731 & 0.1684 & \textbf{0.8807} \\
\bottomrule
\end{tabular}
}
\caption{ACC, ASR, and SIA for BNA, compared with those of NC, I-BAU, ANP, NAD, ARGD, and MCR, with limited amount of clean data on VGGFace2 and CIFAR-10. Both datasets are poisoned by the BadNet attack.}
\label{tab:mitigation_with_few_clean_images}
\end{table}


\subsection{Backdoor Mitigation Results}
\label{subsec:exp_results_main}

In Tab. \ref{tab:comparison}, we show the ASR, ACC, and SIA for BNA compared with the other six methods (which are all DNN tuning-based) for attacks on CIFAR-10. Each metric is averaged over the five attacks created for each trigger type and attack setting, with the highest ACC and SIA, and the lowest ASR in bold. 
We found that these tuning-based methods are sensitive to the choices of hyper-parameters, such as the learning rate. Hence, for these methods, we optimize the hyper-parameter values to show the \textit{best} results for these methods in Tab.~\ref{tab:comparison}.
Although these tuning-based methods (except for MCR) can effectively deactivate backdoor attacks (\ie, significantly reduce ASRs), there is a clear drop (3\%-20\%) in both ACC and SIA, compared with those for the vanilla DNN (the first row of Tab.~\ref{tab:comparison}). This is possibly due to tuning many DNN parameters using very limited data.
(Note that BNA mitigation uses much less clean labelled data than what was reported for these other methods in their
original papers.)
Though MCR can effectively deactivate most of the backdoor attacks, excluding the global pattern CB and $l_2$ inv, it fails to infer the true source classes for the backdoor-triggered instances.
For ANP with neuron pruning, the performance is acceptable only for A2O with the BadNet trigger. One possible reason is that invisible, perturbation-based triggers affect most neurons only moderately (which is also discussed in \cite{UniBD}); thus, pruning a small number of neurons cannot mitigate the attack. In contrast, our method successfully mitigates all these backdoor attacks (with generally the best ACC and ASR compared with the others) regardless of the trigger type and attack setting. Notably, {\it since the purpose of BNA's divergence minimization is to maximize the SIA}, it unsurprisingly achieves the best SIA with a clear margin over all other methods, in all cases (the corresponding distribution divergnces are shown in Tab.~\ref{tab:distribution_divergences} in Apdx.~\ref{sec:distribution_div}). 
We also tune the poisoning ratio and perturbation size used in A2O CB attacks, and the performance for BNA slightly declines as the attack is strengthened, as shown in Tab.~\ref{tab:bd_poison_ratio_pert_size}. However, it still outperforms the other methods (see Tab.~\ref{tab:bd_poison_ratio_pert_size_RW} in Apdx.~\ref{sec:bd_poison_ratio_pert_size}). 

\begin{table}[h!]
\centering
\scriptsize
\begin{tabular}{c|ccccc|ccccc}
\toprule
\hline
& \multicolumn{5}{c|}{Number of poisoned instances per class} & \multicolumn{5}{c}{Perturbation size (*255)} \\
& 50 & 100 & 150 & 200 & 250 & 2 & 3 & 4 & 5 & 6 \\
\hline  
ACC & 0.9112 & 0.9094 & 0.9098 & 0.9102 & 0.9015 & 0.9094 & 0.9041 & 0.9079 & 0.8992 & 0.8912 \\ 
ASR & 0.0095 & 0.0141 & 0.0121 & 0.0170 & 0.0090 & 0.0141 & 0.0395 & 0.0222 & 0.0109 & 0.0388 \\ 
SIA & 0.8851 & 0.8837 & 0.8728 & 0.8840 & 0.8662 & 0.8851 & 0.8783 & 0.8711 & 0.8814 & 0.8435 \\
\hline
\bottomrule
\end{tabular}
\caption{ACC, ASR, and SIA for BNA as a function of (1) the number of poisoned instances injected into the training set; (2) the perturbation size under all-to-one CB attack.}
\label{tab:bd_poison_ratio_pert_size}
\end{table}

For the CL attack, we poison half (2500) of the target-class training samples to achieve an effective attack (which is stronger than in the original paper \citep{CL}), as shown in Tab.~\ref{tab:mitigation_on_CL_attack}. Although NAD and ARGD effectively deactivate the attack, both ACC and ASR drop significantly. For NC, ANP, and MCR, the ACC after mitigation is almost the same as the ACC before mitigation, but the ASR is still high. For ANP, nearly half of the backdoor-trigger images are unimpeded by the mitigation system. The attack is still effective after MCR is deployed. I-BAU does not perform well in mitigating the CL attack -- the mitigated model fails to correctly classify most of the clean test images, but still recognizes half of the backdoor-trigger images to the target class. By contrast, BNA decreases the ACC by only a small amount,
reduces ASR to around 6\%, and correctly classifies 80\% of the backdoor-trigger images.

Results of BNA on other datasets are shown in Tab. ~\ref{tab:mitigation_with_few_clean_images} and \ref{tab:other datasets}. 
We first train a DNN on the VGGFace2 dataset poisoned by the BadNet attack.
As shown in Tab. ~\ref{tab:mitigation_with_few_clean_images}, the BadNet attack is effective, with a high ASR and a nearly unchanged ACC. (The ACC for the DNN trained on the clean VGGFace2 dataset is 0.9211.) We then apply BNA, NC, and I-BAU on the poisoned DNN.\footnote{We did not evaluate the performance of ANP, NAD, and ARGD on VGGFace2, since these references do not provide the architecture of VGG-16 that fits their respective mitigation system. Although MCR provides the architecture of VGG-16, it is different from the one provided by PyTorch. Therefore we cannot load the pre-trained weights and make a fair comparison.} For all mitigation methods, we only preserve 10 clean images per class since there is a severely limited number of samples for VGGFace2. BNA effectively reduces the ASR and yields a high SIA, outperforming the other mitigation methods. We will thoroughly discuss the impact of the number of clean images possessed by the defender in Sec.~\ref{subsec:mitigation_with_limited_clean_data}.
The ACC for DNNs trained without attack for GTSRB, CIFAR-100, ImageNette, and TinyImageNet are 0.9567, 0.6926, 0.8726, and 0.5224, respectively; while ACC, ASR, and SIA for attacked DNNs are shown in the row ``Vanilla'' in Tab.~\ref{tab:other datasets}, which demonstrate that all the attacks are effective. We apply BNA on the poisoned DNNs, with the same settings as for CIFAR-10, which significantly reduces ASR (to less than 1.3\% in all cases), with uniformly high SIA and ACC.

We also evaluated the performance of our BNA against all-to-one backdoor attacks that utilize more complex global backdoor patterns, such as the blended backdoor attack \citep{Targeted-Backdoor} and Sinusoidal Signal backdoor attack (SIG) \citep{SIG}. 
Details of the attack configurations can be found in Apdx.~\ref{sec:attack_settings}.
The performance of our BNA as well as other mitigation methods are shown in Tab.~\ref{tab:mitigation_on_Blend_and_SIG_attack}.
The results demonstrate the effectiveness of our BNA mitigation method even when dealing with complicated backdoor patterns. 
Compared with the other methods\footnote{We weren't able to reproduce the results reported in the published papers describing these methods due to different defense settings -- in our experiments, the defender possesses far fewer clean samples.}, 
our BNA significantly reduces the ASRs and produces relatively satisfactory SIAs, while maintaining ACCs that are competitive with pre-mitigation figures. 

\begin{table}[h!]
\setlength{\tabcolsep}{4pt}
\centering
\resizebox{0.99\textwidth}{!}{
\begin{tabular}{ccccccccccccc}
\toprule
\hline
\multirow{2}{*}{\makecell[c]{Trigger \\ type}} &  & \multicolumn{5}{c}{GTSRB} & \multicolumn{4}{c}{CIFAR-100} & TinyImageNet & ImageNette \\
\cline{3-7} \cline{8-11}  \cline{12-12} \cline{13-13}
&  & BadNet & CB & $l_0$ inv & $l_2$ inv & WaNet & BadNet & CB & $l_0$ inv & $l_2$ inv & BadNet & BadNet \\
\hline
\multirow{3}{*}{\makecell[c]{Vanilla}}
& ACC& 0.9517 & 0.9556 & 0.9531 & 0.9521 & 0.9408 & 0.6796 & 0.6917 & 0.6863 & 0.6804 & 0.5192 & 0.8626 \\
& ASR & 1.0000 & 1.0000 & 1.0000 & 0.9794 & 0.9000 & 0.9037 & 0.9169 & 0.9935 & 0.9097 & 0.8058 & 0.9144 \\
& SIA & 0.0000 & 0.0000 & 0.0000 & 0.0169 & 0.0905 & 0.0781 & 0.0646 & 0.0063 & 0.0707 & 0.1134 & 0.0771 \\
\hline
\multirow{3}{*}{BNA}
& ACC & 0.9491 & 0.9548 & 0.9505 & 0.9500 & 0.9404 & 0.6770 & 0.6863 & 0.6858 & 0.6787 & 0.5178 & 0.7941 \\
& ASR & 0.0000 & 0.0000 & 0.0122 & 0.0001 & 0.0041 & 0.0002 & 0.0524 & 0.0062 & 0.0016 & 0.0043 & 0.0016 \\
& SIA & 0.9312 & 0.9454 & 0.9330 & 0.8945 & 0.9338 & 0.6526 & 0.5880 & 0.6169 & 0.5224 & 0.4965 & 0.7940 \\
\hline
\bottomrule
\end{tabular}
}
\caption{ACC, ASR, and SIA for BNA against all-to-one attacks on CIFAR-100, GTSRB, ImageNette, and TinyImageNet datasets.}
\label{tab:other datasets}
\end{table}

\begin{table}[h!]
\centering
\scriptsize
\begin{tabular}{ccccccccc}
\toprule
\hline
&  & Vanilla & NC & I-BAU & ANP & NAD & ARGD & BNA \\
\hline
\multirow{3}{*}{Blend}
& ACC & 0.9264 & 0.8132 & 0.7932 & 0.8667 & 0.8221 & 0.4856 & \textbf{0.8942} \\ 
& ASR & 0.9731 & \textbf{0.0488} & 0.7419 & 0.5119 & 0.0521 & 0.0782 & 0.1283 \\ 
& SIA & 0.0254 & 0.5593 & 0.1427 & 0.3369 & 0.6003 & 0.4181 & \textbf{0.6252} \\ 
\hline
\multirow{3}{*}{SIG}
& ACC & 0.9266 & 0.8234 & 0.5696 & 0.8251 & 0.7794 & 0.4233 & \textbf{0.8716} \\ 
& ASR & 0.9991 & 0.1414 & 0.2594 & 0.9451 & 0.3223 & 0.0988 & \textbf{0.0158} \\ 
& SIA & 0.0008 & 0.2980 & 0.1319 & 0.0383 & 0.2503 & 0.3271 & \textbf{0.3357} \\ 
\hline
\bottomrule
\end{tabular}
\caption{ACC, ASR, and SIA for BNA, compared with those of NC, NAD, I-BAU, ANP, and ARGD, against the ResNet-18 trained on the CIFAR-10 dataset poisoned by the all-to-one Blend and SIG backdoor attacks.}
\label{tab:mitigation_on_Blend_and_SIG_attack}
\end{table}

\begin{table}[h!]
\centering
\resizebox{.99\textwidth}{!}{
\begin{tabular}{cccccccccccc}
\toprule
\hline
$\tau$ ($\Delta b$=0.1) & 10 & 100 & 200 & 300 & 400 & 500 & 600 & 700 & 800 & 900 & 1000 \\
ACC & 0.9024 & 0.9025 & 0.9022 & 0.9019 & 0.9024 & 0.9019 & 0.9022 & 0.9018 & 0.9017 & 0.9015 & 0.9021 \\
ASR & 0.0257 & 0.022 & 0.0206 & 0.0207 & 0.0214 & 0.0202 & 0.0212 & 0.0209 & 0.0207 & 0.0201 & 0.0204 \\
SIA & 0.8744 & 0.8758 & 0.8774 & 0.8768 & 0.8768 & 0.8775 & 0.8773 & 0.8768 & 0.8772 & 0.8778 & 0.8777 \\
\hline
$\Delta b$ ($\tau$=150) & 0.1 & 0.11 & 0.12 & 0.13 & 0.14 & 0.15 & 0.16 & 0.17 & 0.18 & 0.19 & 0.2 \\
ACC & 0.9028 & 0.9018 & 0.9019 & 0.9023 & 0.9019 & 0.9023 & 0.9019 & 0.9019 & 0.9022 & 0.9022 & 0.9022 \\
ASR & 0.0197 & 0.0202 & 0.0199 & 0.0204 & 0.0199 & 0.0204 & 0.0198 & 0.0206 & 0.0206 & 0.0207 & 0.0212 \\
SIA & 0.8799 & 0.8775 & 0.8779 & 0.8773 & 0.8772 & 0.8779 & 0.8773 & 0.8767 & 0.8779 & 0.8769 & 0.8764 \\
\hline
\bottomrule
\end{tabular}
}
\caption{ACC, ASR, and SIA for BNA as a function of scale factor and bin size on ResNet-18 trained on CIFAR-10 poisoned by all-to-one BadNet attack. }
\label{tab:t&bin_size}
\end{table}

\subsection{The ``detection-before-mitigation'' scenario}\label{subsec:detection_before_mitigation}
As discussed in Sec.~\ref{subsec:backdoor_mitigation_problem}, BNA performs backdoor mitigation only after the model has been detected as backdoor-poisoned.
To justify the ``detection-before-mitigation scenario'', we first apply the mitigation methods that do not involve a detection system (\ie, I-BAU, ANP, NAD, ARGD, and MCR) on a ResNet-18 trained on the attack-free CIFAR-10 dataset. The resulting (absolute) drop in ACC is shown in column ``clean'' in Tab.~\ref{tab:mitigation_on_clean_model}. 
ANP has the largest impact on ACC -- the ACC drops by 0.1877 after mitigation. I-BAU and ARGD respectively decrease the ACC by 0.0684 and 0.0343. NAD and MCR keep the ACC almost as high as that of the vanilla model, but they are not sufficiently effective in terms of SIA when the model is poisoned. NC and BNA detect the backdoor attack before mitigation; thus there is no impact on the ACC for clean classifiers (for which no attack is detected).
We also found reduction in ACCs when applying all mitigation methods for a ResNet-18 model trained on CIFAR-10 poisoned by an all-to-one BadNet attack in column ``BadNet''. All the other methods decrease ACC by more than 0.03, while our method has little impact on ACC.

\begin{table}[h!]
\setlength{\tabcolsep}{4pt}
\centering
\resizebox{.99\textwidth}{!}{
\begin{tabular}{cccccccccccccc}
\toprule
\hline
\multirow{2}{*}{\makecell[c]{Trigger \\ type}} &  &  \multicolumn{2}{c}{BadNet} & \multicolumn{2}{c}{CB} & \multicolumn{2}{c}{$l_0$ inv} & \multicolumn{2}{c}{$l_2$ inv} & \multicolumn{2}{c}{SP} & \multicolumn{2}{c}{WaNet} \\
\cline{3-14}
&  & A2O & A2A & A2O & A2A & A2O & A2A & A2O & A2A & A2O & A2A & A2O & A2A \\ 
\hline
\multirow{2}{*}{BNA} 
& FPR & \textbf{0.1390} & \textbf{0.0606} & \textbf{0.1144} & \textbf{0.1092} & \textbf{0.1413} & \textbf{0.0600} & 0.1976 & \textbf{0.1027} & \textbf{0.1323} & \textbf{0.0656} & \textbf{0.1406} & \textbf{0.0865} \\
& TPR & \textbf{0.9872} & \textbf{0.9508} & \textbf{0.9872} & \textbf{0.9682} & 0.9967 & \textbf{0.9873} & \textbf{0.9958} & \textbf{0.9793} & \textbf{0.9894} & \textbf{0.9294} & \textbf{0.9959} & \textbf{0.9248} \\
\hline
\multirow{2}{*}{STRIP}
& FPR & 0.15 & 0.15 & 0.15 & 0.15 & 0.15 & 0.15 & \textbf{0.15} & 0.15 & 0.15 & 0.15 & 0.15 & 0.15 \\ 
& TPR & 0.9638 & 0.5147 & 0.6802 & 0.2089 & \textbf{0.9995} & 0.1053 & 0.9924 & 0.5272 & 0.8522 & 0.3123 & 0.0202 & 0.0411 \\
\hline
\bottomrule
\end{tabular}
}
\caption{TPR and FPR for BNA, compared with STRIP, against all attacks created on CIFAR-10.}
\label{tab:test_time_detection}
\end{table}

\begin{table}[h!]
\centering
\resizebox{0.99\textwidth}{!}{
\begin{tabular}{cccccccccccccc}
\toprule
\hline
\multicolumn{2}{c}{I-BAU} & \multicolumn{2}{c}{ANP} & \multicolumn{2}{c}{NAD} & \multicolumn{2}{c}{ARGD} & \multicolumn{2}{c}{MCR} & \multicolumn{2}{c}{NC} & \multicolumn{2}{c}{BNA} \\
\cline{1-14}
Clean & BadNet & Clean & BadNet & Clean & BadNet & Clean & BadNet & Clean & BadNet & Clean & BadNet & Clean & BadNet \\
0.0684 & 0.0622 & 0.1877 & 0.0478 & 0.0064 & 0.0308 & 0.0343 & 0.0433 & 0.0038 & 0.0371 & NA & 0.0325 & NA & 0.0090 \\
\bottomrule
\end{tabular}
}
\caption{Drop in ACC when applying I-BAU, ANP, NAD, ARGD, NC and BNA on ResNet-18 trained on the clean (attack-free) CIFAR-10 dataset and trained on CIFAR-10 poisoned by the all-to-one BadNet attack. }
\label{tab:mitigation_on_clean_model}
\end{table}

\subsection{Test-Time backdoor-trigger instance detection}
Different from other tuning-based backdoor mitigation approaches, our BNA can also detect backdoor-trigger instances at test-time, as described in Sec.~\ref{subsec:backdoor_mitigation_method} and shown in Fig.~\ref{fig:procedure}. Here, we evaluate accuracy of our test-time detector compared with a state-of-the-art detector named STRIP \citep{STRIP}. For any input image during inference, STRIP blends it with clean images possessed by the defender. The blended image is fed into the poisoned DNN, with an entropy calculated on the output posteriors. If the entropy is lower than a prescribed detection threshold, the input is deemed to be embedded with the trigger. Here, we set the detection threshold to achieve 15\% FPR for STRIP, a choice which achieves a generally good trade-off between TPR and FPR. In contrast, BNA does not require setting a detection threshold.
In Tab.~\ref{tab:test_time_detection}, we show the True Positive Rate (TPR, \ie, the fraction of backdoor-trigger images correctly detected) and the False Positive Rate (FPR, \ie, the fraction of clean test images from the backdoor target class(es) that are falsely detected) for both methods. Although STRIP performs well on A2O attacks for some trigger types, \eg, BadNet, $l_0$ inv, and $l_2$ inv, its TPR drops drastically on attacks using human-imperceptible triggers, especially the WaNet attacks.
Moreover, it does not perform well on all A2A attacks, with a largest TPR of only 0.5272.
By contrast, BNA is effective for all these attacks -- it detects almost all the backdoor-trigger images, with FPRs comparable to STRIP.

\subsection{Mitigation performance against adaptive attacks}
A recent backdoor attack proposed by \cite{WB} minimizes a metric similar to that used by our BNA defense, in order to achieve better stealthiness. This attack can be viewed as an adaptive attack against our mitigation defense since the trained classifier will be more sensitive to even a smaller distribution divergence than for ordinary backdoor attacks. Nevertheless, our method successfully mitigates this attack. In our experiment on CIFAR-10, the average distribution total variation divergence over all neurons is reduced from 8067 to 2789. Accordingly, the ACC/ASR before and after mitigation are 0.9162/0.9978 and 0.8906/0.0072, respectively, with an SIA of 0.8496.

\subsection{Mitigation with a limited amount of clean data}\label{subsec:mitigation_with_limited_clean_data}
{\it Why tuning-based methods like NC cannot achieve SIAs as high as BNA 
(which does not alter the DNN's parameters) ?} Note that NC tunes the classifier using instances embedded with the estimated trigger but without label flipping. This is equivalent to minimizing the divergence between internal activation distributions for clean and triggered instances (see Tab.~\ref{tab:distribution_divergences} in Apdx.~\ref{sec:distribution_div}), but by altering the DNN's parameters. Even for an optimal (zero) divergence, the best achievable SIA of NC is still upper-bounded by the ACC of the classifier after tuning, which usually {\it drops} due to the \textbf{data insufficiency}. By contrast, the reference distribution for BNA's divergence minimization is obtained by feeding clean instances to the poisoned classifier {\it without} changing its parameters; thus, it is a ``better'' reference with a higher upper-bound ACC.

For the main experiments on CIFAR-10, we preserve 100 clean test images for all mitigation methods. In other words, all mitigation methods, excluding our BNA, tune the (around 11 million) trainable parameters of the poisoned ResNet-18 based only on 1000 clean labelled images (2\% of the training set) for a few epochs. 
Our method is light-weight, since it only updates the (less than 10 thousand) non-trainable parameters (\ie, mean and standard deviation).
Note that all the mitigation methods use more clean images in their original papers than in the experiments reported herein. For example, NC and NAD respectively chose 10\% and 5\% of the clean training samples for mitigation.
For all mitigation methods, excluding our BNA, the \textbf{insufficiency} of clean labelled data reduces the ACC by at most 10\%. The SIA is upper-bounded by the ACC after mitigation and is also affected by data insufficiency. 
This is also verified in \cite{NAD}, where they varied the number of clean images from 0\% to 20\% of the clean training samples, with the performance of NAD significantly degraded as the number of clean samples decreases. 
Our BNA only aligns internal layer distributions without affecting the trainable parameters, and  thus is more robust  when the amount of clean samples is limited.

To further demonstrate the impact of the number of clean samples on mitigation performance, for the ResNet-18 trained on CIFAR-10 poisoned by the BadNet attack, we reduce the number of clean images used by the defender to just 10 images per class. The corresponding performance of all mitigation methods is shown in Tab.~\ref{tab:mitigation_with_few_clean_images}. Although ANP and ARGD effectively de-activate the backdoor attack, both ACC and SIA dramatically decrease. For NC, I-BAU, NAD, and MCR, the ACC changes a little, but the attack is still effective, especially for NAD and MCR. However, our BNA is still effective, with the ASR  less than 2\% and both ACC and SIA  comparable to the ACC before mitigation. 

\textbf{Data insufficiency} is a common phenomenon in real-world applications. For example, we use a subset of VGGFace2, which consists of 18 identities, each of which has 450 training face images and 100 test face images. So, VGGFace2 is much smaller than other benchmark datasets such as CIFAR-10. We only assign 10 clean images per class for the defender. The results are shown in Tab.~\ref{tab:mitigation_with_few_clean_images}. On this high-resolution and insufficient dataset, both NC and I-BAU fail to de-activate the BadNet attack. In contrast, our BNA successfully reduces the ASR to 0.46\% and has ACC and SIA about 89\%.

\section*{Conclusion}
In this paper, we revealed an activation distribution alteration property for backdoor attacks. 
We theoretically proved that by correcting such alteration, backdoor trigger instances will be correctly classified to their original source classes. 
Accordingly, we proposed a post-training backdoor mitigation approach to align distributions of clean and backdoor-trigger samples through simple transformations, \textit{without changing millions trainable parameters} of the classifier, which outperformed methods that use DNN fine-tuning. 
The proposed method is \textit{robust} especially when there is \textit{limited amount} of clean data available to the defender, compared with parameter-tuning based methods.
Besides, the proposed method is \textit{flexible} to be integrated with existing detection systems. 
Moreover, our method can detect instances with the trigger during inference.

\section*{Ethics Statement}\label{ethics_statement}
The main purpose of this research is to understand the behavior of deep learning systems facing malicious activities, and enhance their safety level. The backdoor attack considered in this paper is well-known, with open-sourced implementation code. Thus, publication of this paper will be beneficial to the community in defending against backdoor attacks. The code of our defense will be released if the paper is accepted.

\subsection*{Acknowledgments}
This research was supported in part by NSF SBIR grant 2132294.

\section*{Appendix}

\section{Proof of Theorems in the Main Paper}\label{sec:proof}

\subsection{Derivation of Eq. (\ref{eq:mean_var_updated})}\label{subsec:proof_optimal_solution}

Here, we provide the derivation showing that $m^{\ast}_j$ and $v^{\ast}_j$ in Eq. (\ref{eq:mean_var_updated}) are the solutions to:
\begin{align}
	{\mathbb E}[a^{\ast}_j({\bf X}_b)] &= {\mathbb E}[a_j({\bf X})|Y=-1] \label{eq:equal_mean}\\
	\Var[a^{\ast}_j({\bf X}_b)] &= \Var[a_j({\bf X})|Y=-1] \label{eq:equal_var}
\end{align}
Based on Eq. (\ref{eq:activation}), the above equations can be expanded as  follows:
\begin{align}
	{\mathbb E}[\frac{\vw_j^\top {\bf X}_b - m^{\ast}_j}{\sqrt{v^{\ast}_j}} \gamma_j + \beta_j] &= {\mathbb E}[\frac{\vw_j^\top {\bf X} - m_j}{\sqrt{v_j}} \gamma_j + \beta_j | Y=-1] \label{eq:optimal_mean_0}\\
	\Var[\frac{\vw_j^\top {\bf X}_b - m^{\ast}_j}{\sqrt{v^{\ast}_j}} \gamma_j + \beta_j] &= \Var[\frac{\vw_j^\top {\bf X} - m_j}{\sqrt{v_j}} \gamma_j + \beta_j | Y=-1] \label{eq:optimal_var_0}
\end{align}
We first solve Eq. (\ref{eq:optimal_var_0}) for $({\bf X}|Y=-1) \sim \mathcal{N}(-\vmu, \sigma^{2}\mI)$ and ${\bf X}_b \sim \mathcal{N}(\vmu_b, \sigma_b^{2}\mI)$, which leads to:
\begin{align}\label{eq:optimal_var_1}
	v^{\ast}_j=\frac{\sigma_b}{\sigma} v_j
\end{align}
By substituting Eq. (\ref{eq:optimal_var_1}) into Eq. (\ref{eq:optimal_mean_0}), and since $\vmu_b=-\vmu+\bm{\epsilon}$, we get the following:
\begin{align*}
	m^{\ast}_j &= \sqrt{\frac{v^{\ast}_j}{v_j}} (m_j - \vw_j^\top\vmu) + \vw_j^\top\vmu_b\\
	&=\frac{\sigma_b}{\sigma} m_j + (\frac{\sigma_b}{\sigma} - 1) \vw_j^\top \vmu + \vw_j^\top \bm{\epsilon}.
\end{align*}

\subsection{Proof of Theorem \ref{thm:main}}\label{subsec:proof_main}

\begin{proof}
First, let's specify the following vector/matrix representations that will be used throughout this proof:
\begin{align*}
	\mW = [\vw_1, \cdots, \vw_J]^\top \in {\mathbb R}^{J\times d}
\end{align*}
\begin{align*}
	\mV = \begin{bmatrix}
		v_1 & \cdots & 0 \\
		 \vdots & \ddots & \vdots\\
		 0 & \cdots & v_J
	\end{bmatrix}\in{\mathbb R}^{J\times J} \quad\quad
	\hat{\mV}(\alpha) = \begin{bmatrix}
		\hat{v}_1(\alpha) & \cdots & 0\\
		\vdots & \ddots & \vdots\\
		0 & \cdots & \hat{v}_J(\alpha)
	\end{bmatrix}\in{\mathbb R}^{J\times J}
\end{align*}
\begin{align*}
	\vm = \begin{bmatrix}
		m_1\\
		\vdots\\
		m_J
	\end{bmatrix}\in{\mathbb R}^{J} \,\,
	\hat{\vm}(\alpha) = \begin{bmatrix}
		\hat{m}_1(\alpha)\\
		\vdots\\
		\hat{m}_J(\alpha)
	\end{bmatrix}\in{\mathbb R}^{J} \,\,
	{\bm \Gamma} = \begin{bmatrix}
		\gamma_1 & \cdots & 0 \\
		\vdots & \ddots & \vdots\\
		0 & \cdots & \gamma_J
	\end{bmatrix}\in{\mathbb R}^{J\times J} \,\,
	\bm{\beta} = \begin{bmatrix}
		\bm{\beta}_1\\
		\vdots\\
		\bm{\beta}_J
	\end{bmatrix}\in{\mathbb R}^{J}
\end{align*}
\begin{align*}
	\va(\cdot) = \begin{bmatrix}
		a_1(\cdot)\\
		\vdots\\
		a_J(\cdot)
	\end{bmatrix}\in{\mathbb R}^{J} \quad
	\hat{\va}(\cdot|\alpha) = \begin{bmatrix}
		\hat{a}_1(\cdot|\alpha)\\
		\vdots\\
		\hat{a}_J(\cdot|\alpha)
	\end{bmatrix}\in{\mathbb R}^{J} \quad
	\va^{\ast}(\cdot) = \begin{bmatrix}
		a_1^{\ast}(\cdot)\\
		\vdots\\
		a_J^{\ast}(\cdot)
	\end{bmatrix}\in{\mathbb R}^{J}
\end{align*}

Let ${\bf X}_-=({\bf X}|Y=-1) \sim \mathcal{N}(-\vmu, \sigma^{2}\mI)$ denote a random instance from source class `$-1$' for simplicity. Let $\vu=\vu_- - \vu_+$ with $\vu_-$ and $\vu_+$ being the weight vectors associated with the node for class `$-1$' and the node for class `$+1$' respectively. Then, it is easy to see that:
\begin{align*}
	\hat{\va}({\bf X}_b|\alpha)\big|_{\alpha=1} = \va({\bf X}_b) \quad{\text{and}}\quad \hat{\va}({\bf X}_b|\alpha)\big|_{\alpha=0} = \va^{\ast}({\bf X}_b),
\end{align*}
and taking one step further by setting $\alpha=1$, we have the following:
\begin{align}
	P[g_-({\bf X}_b|\alpha) > g_+({\bf X}_b|\alpha)\big|\alpha=1] &= P[\vu^\top \hat{\va}({\bf X}_b|\alpha) > 0 \big|\alpha=1] \label{eq:set2one}\\
	&= P[\vu^\top \va({\bf X}_b) > 0] \nonumber\\
	&= P[f_-({\bf X}_b) > f_+({\bf X}_b)] \nonumber\\
	&\leq 1-\psi. \label{eq:psi_successful}
\end{align}
This is, when $\alpha=1$, the classifier is not modified at all; thus the SIA will be no larger than $1-\psi$ since the attack is $\psi$-successful (see Definition \ref{def:psi_successful}). On the other hand, by setting $\alpha=0$, we will have the following:
\begin{align}
	P[g_-({\bf X}_b|\alpha) > g_+({\bf X}_b|\alpha)\big|\alpha=0] &= P[\vu^\top \hat{\va}({\bf X}_b|\alpha) > 0 \big|\alpha=0] \nonumber\\
	&= P[\vu^\top \va^{\ast}({\bf X}_b) > 0] \nonumber\\
	&= P[f_-({\bf X}_-) > f_+({\bf X}_-)] \label{eq:equal_dist}\\
	&\geq 1 - \eta. \label{eq:eta_erroneous}
\end{align}
That is, when $\alpha=0$, the distribution shift will be fully recovered, such that SIA is equally high as the accuracy of the source class. Recall that Eq. (\ref{eq:equal_dist}) is due to Eq. (\ref{eq:equal_mean}) and Eq. (\ref{eq:equal_var}). The inequality (\ref{eq:eta_erroneous}) is because the classifier specified by $f_-$ and $f_+$ is assumed $\eta$-erroneous (see Definition \ref{def:eta_erroneous}). Here, we prove the theorem by showing that the partial derivative of $P[g_-({\bf X}_b|\alpha) > g_+({\bf X}_b|\alpha)]$ over $\alpha$ is strictly {\it negative} when $\sigma_b\leq\sigma$ (\ie, triggered instances have smaller standard deviation than clean instances, which is generally true). To achieve this, we notice that
\begin{align*}
	\vu^\top \hat{\va}({\bf X}_b|\alpha) = \vu^\top \hat{\mV}(\alpha)^{-\frac{1}{2}} {\bm \Gamma} (\mW {\bf X}_b - \hat{\vm}(\alpha)) + \vu^\top \bm{\beta}
\end{align*}
follows a Gaussian distribution with
\begin{align}
	{\mathbb E}[\vu^\top \hat{\va}({\bf X}_b|\alpha)] &= \vu^\top \hat{\mV}(\alpha)^{-\frac{1}{2}} {\bm \Gamma} (-\mW \vmu + \mW \bm{\epsilon} - \hat{\vm}(\alpha)) + \vu^\top \bm{\beta} \label{eq:expectation_after}\\
	\Var[\vu^\top \hat{\va}({\bf X}_b|\alpha)] &= \sigma_b^{2} || \mW^\top {\bm \Gamma} \hat{\mV}(\alpha)^{-\frac{1}{2}} \vu||_2^2 \label{eq:var_after}
\end{align}
We also notice that for source class instances ${\bf X}_-$
\begin{align*}
	\vu^\top \va({\bf X}_-) = \vu^\top \mV^{-\frac{1}{2}} {\bm \Gamma} (\mW {\bf X}_- - \vm) + \vu^\top \bm{\beta}
\end{align*}
follows a Gaussian distribution with
\begin{align}
	{\mathbb E}[\vu^\top \va({\bf X}_-)] &= \vu^\top \mV^{-\frac{1}{2}} {\bm \Gamma} (-\mW \vmu - \vm) + \vu^\top \bm{\beta} \label{eq:expectation_before}\\
	\Var[\vu^\top \va({\bf X}_-)] &= \sigma^{2} || \mW^\top {\bm \Gamma} \mV^{-\frac{1}{2}} \vu||_2^2
\end{align}
Then we have
\begin{align}
	P[\vu^\top \hat{\va}({\bf X}_b|\alpha) > 0] &= 1 - \mPhi(-\frac{{\mathbb E}[\vu^\top \hat{\va}({\bf X}_b|\alpha)]}{\sqrt{\Var[\vu^\top \hat{\va}({\bf X}_b|\alpha)]}}) \label{eq:phi_function_modified}\\
	P[\vu^\top \va({\bf X}_-) > 0] &= 1 - \mPhi(-\frac{{\mathbb E}[\vu^\top \va({\bf X}_-)]}{\sqrt{\Var[\vu^\top \va({\bf X}_-)]}}) \label{eq:phi_function}
\end{align}
where $\Phi$ is the cumulative distribution function of standard Gaussian. Now let's consider Eq. (\ref{eq:phi_function}) first. Since $\eta<\frac{1}{2}$ as we have reasonably assumed (otherwise the classifier may be worse than a random guess), and also according to Eq. (\ref{eq:eta_erroneous}), we have
\begin{align*}
	P[\vu^\top \va({\bf X}_-) > 0] = P[f_-({\bf X}_-) > f_+({\bf X}_-)] > \frac{1}{2}
\end{align*}
Thus, based on Eq. (\ref{eq:phi_function}) and Eq. (\ref{eq:expectation_before}), we get
\begin{align}
	\vu^\top \mV^{-\frac{1}{2}} {\bm \Gamma} (-\mW \vmu - \vm) + \vu^\top \bm{\beta} > 0 \label{eq:assumption_eta}
\end{align}
Next, let us focus on Eq. (\ref{eq:phi_function_modified}). Again, we set $\alpha=1$. Based on (\ref{eq:set2one})-(\ref{eq:psi_successful}) and the reasonable assumption that $\psi>\frac{1}{2}$ (otherwise the attack is not deemed successful since the success rate will be even lower than the accuracy on clean instances), we have
\begin{align*}
	P[\vu^\top \hat{\va}({\bf X}_b|\alpha) > 0 \big|\alpha=1] = P[f_-({\bf X}_b) > f_+({\bf X}_b)] < \frac{1}{2}
\end{align*}
Then, based on Eq. (\ref{eq:phi_function_modified}) and Eq. (\ref{eq:expectation_after}), we get
\begin{align}
	\vu^\top \mV^{-\frac{1}{2}} {\bm \Gamma} (-\mW \vmu + \mW \bm{\epsilon} - \vm) + \vu^\top \bm{\beta} < 0 \label{eq:assumption_psi}
\end{align}
Subtracting Eq. (\ref{eq:assumption_psi}) from Eq. (\ref{eq:assumption_eta}) we get:
\begin{align}\label{eq:clue}
	-\vu^\top \mV^{-\frac{1}{2}} {\bm \Gamma} \mW \bm{\epsilon} > 0
\end{align}
Based on Eq. (\ref{eq:phi_function_modified}), we also have
\begin{align}\label{eq:derivative}
	\frac{\partial P[g_-({\bf X}_b|\alpha) > g_+({\bf X}_b|\alpha)]}{\partial \alpha} = \frac{\partial\frac{{\mathbb E}[\vu^\top \hat{\va}({\bf X}_b|\alpha)]}{\sqrt{\Var[\vu^\top \hat{\va}({\bf X}_b|\alpha)]}}}{\partial \alpha} \cdot 
	\phi(-\frac{{\mathbb E}[\vu^\top \hat{\va}({\bf X}_b|\alpha)]}{\sqrt{\Var[\vu^\top \hat{\va}({\bf X}_b|\alpha)]}})
\end{align}
where $\phi$ is the probability density function (PDF) for standard normal distribution. Based on Eq. (\ref{eq:expectation_after}), Eq. (\ref{eq:var_after}), and Eq. (\ref{eq:mean_var_updated}), we have
\begin{align*}
	&\frac{{\mathbb E}[\vu^\top \hat{\va}({\bf X}_b|\alpha)]}{\sqrt{\Var[\vu^\top \hat{\va}({\bf X}_b|\alpha)]}}\\
	= &\frac{\vu^\top \mV^{-\frac{1}{2}} {\bm \Gamma} [-\mW \vmu + \mW \bm{\epsilon} - (\alpha\vm + (1-\alpha)\vm^{\ast})] + (\alpha + (1-\alpha)\frac{\sigma_b}{\sigma}) \vu^\top \bm{\beta}} {\sigma_{b} || \mW^\top {\bm \Gamma} \mV^{-\frac{1}{2}} \vu||_2}\\
	= &\frac{\alpha \cdot \vu^\top \mV^{-\frac{1}{2}} {\bm \Gamma} [(\frac{\sigma_b}{\sigma} - 1) \vm + (\frac{\sigma_b}{\sigma} - 1) \mW \vmu + \mW \bm{\epsilon}] - \alpha \cdot (\frac{\sigma_b}{\sigma} - 1) \vu^\top \bm{\beta}} {\sigma_{b} || \mW^\top {\bm \Gamma} \mV^{-\frac{1}{2}} \vu||_2} + \text{constant}
\end{align*}
and thus, based on Eq. (\ref{eq:assumption_psi}) and Eq. (\ref{eq:clue})
\begin{align*}
	&\frac{\partial\frac{{\mathbb E}[\vu^\top \hat{\va}({\bf X}_b|\alpha)]}{\sqrt{\Var[\vu^\top \hat{\va}({\bf X}_b|\alpha)]}}}{\partial \alpha}\\
	=& \frac{(\frac{\sigma_b}{\sigma} - 1) [\vu^\top \mV^{-\frac{1}{2}} {\bm \Gamma} (\vm + \mW \vmu - \mW \bm{\epsilon}) - \vu^\top \bm{\beta}] + \frac{\sigma_b}{\sigma} \vu^\top \mV^{-\frac{1}{2}} {\bm \Gamma} \mW \bm{\epsilon}}{\sigma_{b} || \mW^\top {\bm \Gamma} \mV^{-\frac{1}{2}} \vu||_2}\\
	<& 0
\end{align*}
when $\sigma_b\leq\sigma$. Substitute it into Eq. (\ref{eq:derivative}) and given the Gaussian PDF being strictly positive, we have
\begin{align*}
	\frac{\partial P[g_-({\bf X}_b|\alpha) > g_+({\bf X}_b|\alpha)]}{\partial \alpha} < 0
\end{align*}

\end{proof}

\section{Datasets, Training settings, and attack settings}

\subsection{Datasets}\label{sec:datasets}
In the experiments, we show the effectiveness of our proposed backdoor mitigation method on several benchmark datasets including CIFAR-10 (\cite{cifar10}), GTSRB (\cite{GTSRB}), CIFAR-100 (\cite{cifar10}), ImageNette (\cite{imagewang}), and TinyImageNet.
CIFAR-10 dataset contains 60,000 $32\times32$ color images from 10 classes, with 5,000 images per class for training and 1,000 images per class for testing . 
GTSRB has more than 50,000 
traffic sign images with different sizes from 43 classes. Here, we resize all images in GTSRB to $32\times 32$.
CIFAR-100 contains 60,000 $32\times32$ color images evenly from 100 classes, where 500 images per class are used for training, while the others are used for testing. 
ImageNette is a subset of 10 easily classified 
classes from Imagenet\footnote{The 10 classes are tench, English springer, cassette player, chain saw, church, French horn, garbage truck, gas pump, golf ball, and parachute.}, with image size of $256\times256$. For each class, there are around 900 images for training and 400 images for testing.
TinyImageNet is a subset of ImageNet (\cite{ImageNet}).
It contains 100,000 $64\times64$ color images evenly distributed in 200 classes (500 training images and 50 test images for each class). 
VGGFace2 is a large-scale face recognition dataset. It contains more than 3.3 million face images from more than 9000 identities. In our experiments, we use a subset of VGGFace2, which consists of 18 identities, each of which has 550 face images. For each class (identity), 450 images are used for training and 100 images are used for testing. All images are resized to 
$224\times 224$ pixels.

\subsection{Attack Settings}\label{sec:attack_settings}
\begin{figure}[h!]
	\centering
	\begin{subfigure}[]{.2\textwidth}
         \centering
         \includegraphics[width=\textwidth]{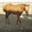}
         \caption{clean}
         \label{fig:clean_visual}
     \end{subfigure}
	\begin{subfigure}[]{.2\textwidth}
		\centering
		\includegraphics[width=\textwidth]{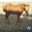}
		 \caption{BadNet}
		 \label{fig:BadNet_visual}
	\end{subfigure}
    \begin{subfigure}[]{.2\textwidth}
         \centering
         \includegraphics[width=\textwidth]{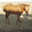}
         \caption{CB}
         \label{fig:CB_visual}
     \end{subfigure}
    \begin{subfigure}[]{.2\textwidth}
         \centering
         \includegraphics[width=\textwidth]{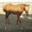}
         \caption{SP}
         \label{fig:SP_visual}
     \end{subfigure}
    \begin{subfigure}[]{.2\textwidth}
         \centering
         \includegraphics[width=\textwidth]{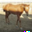}
         \caption{$l_0$ inv}
         \label{fig:l0_visual}
     \end{subfigure}
    \begin{subfigure}[]{.2\textwidth}
         \centering
         \includegraphics[width=\textwidth]{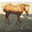}
         \caption{$l_2$ inv}
         \label{fig:l2_visual}
     \end{subfigure}
    \begin{subfigure}[]{.2\textwidth}
         \centering
         \includegraphics[width=\textwidth]{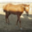}
         \caption{WaNet}
         \label{fig:WaNet_visual}
     \end{subfigure}
	\caption{Example of CIFAR-10 images embedded with the backdoor triggers considered in our experiments.}
	\label{fig:poisoned_images}
\end{figure}

\begin{figure}[h!]
    \centering
    \begin{subfigure}[]{.2\textwidth}
         \centering
         \includegraphics[width=\textwidth]{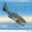}
         \caption{clean}
         \label{fig:clean_2_visual}
     \end{subfigure}
     \begin{subfigure}[]{.2\textwidth}
         \centering
         \includegraphics[width=\textwidth]{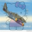}
         \caption{Blend attack}
         \label{fig:Blend_visual}
     \end{subfigure}
     \begin{subfigure}[]{.2\textwidth}
         \centering
         \includegraphics[width=\textwidth]{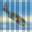}
         \caption{SIG attack}
         \label{fig:SIG_visual}
     \end{subfigure}
    \caption{Example of CIFAR-10 images poisoned by the Blend and SIG backdoor attacks.}
    \label{fig:Blend_and_SIG_images}
\end{figure}

On CIFAR-10, we consider the following triggers:
1) a $3\times3$ random patch (\textbf{BadNet}) with a randomly selected location (fixed for all triggered images for each attack) used in \cite{BadNets}, as visualized in Fig.~\ref{fig:BadNet_visual};
2) an additive perturbation (with size 2/255) resembling a chessboard (\textbf{CB}) used in \cite{TNNLS}, as visualized in Fig.~\ref{fig:CB_visual};
3) a single pixel (\textbf{SP}) perturbed by 75/255 with a randomly selected location (fixed for all triggered images for each attack) used by \cite{SS}, as visualized in Fig.~\ref{fig:SP_visual}; 
4) invisible triggers generated with $l_0$ and $l_2$ norm constraints (\textbf{$l_0$ inv} and \textbf{$l_2$ inv} respectively) proposed by \cite{invisible}, as visualized in Fig.~\ref{fig:l0_visual} and \ref{fig:l2_visual};
5) a warping-based trigger (\textbf{WaNet}) proposed by \cite{WaNet}, as visualized in Fig.~\ref{fig:WaNet_visual};
6) a Hello Kitty trigger with a blend ratio of $\alpha=0.15$ used by \cite{Targeted-Backdoor}, as is visualized in Fig.~\ref{fig:Blend_visual};
7) a sinusoidal signal trigger generated by the horizontal sinusoidal function defined in \cite{SIG} with $\Delta=20$ and $f=6$, as visualized in Fig.~\ref{fig:SIG_visual}.

Attack settings for CIFAR-10 are summarized in Tab.~\ref{tab:attack_config_CIFAR-10}. 
For all-to-one attacks, we arbitrarily choose class 9 as the target class, and embed the backdoor triggers in 100 randomly chosen training samples per class (excluding the target class). 
To achieve similarly effective attacks as for other triggers, we poison 900 images per source class in the all-to-one attack using WaNet.
For all-to-all attacks, we embed the backdoor triggers into 300 images for each class. 
For effective attacks, we poison 800 training images and 1500 training images in the all-to-all attacks using SP and WaNet, respectively.

\begin{figure}[h!]
    \centering
    \begin{subfigure}[]{.2\textwidth}
         \centering
         \includegraphics[width=\textwidth]{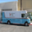}
         \caption{clean}
         \label{fig:target_clean_visual}
     \end{subfigure}
     \begin{subfigure}[]{.2\textwidth}
         \centering
         \includegraphics[width=\textwidth]{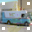}
         \caption{CL attack}
         \label{fig:target_CL_visual}
     \end{subfigure}
    \caption{Example of CIFAR-10 images poisoned by the label-consistent backdoor attack.}
    \label{fig:CL_poisoned_images}
\end{figure}

For CIFAR-10, we also launched the label-consistent (CL) backdoor attack proposed in \citet{CL}, as visualized in Fig.~\ref{fig:target_CL_visual}. We followed the same settings as reported in their paper. Specifically, we used Projected Gradient Descent (PGD) to generate adversarial perturbations bounded by a $L_{\inf}$ maximum perturbation of $\epsilon = 16$. A $3\times 3$ white square is embedded at the four corners of the perturbed images. Half (2500) of the target class training images are poisoned by the CL attack. All non-target class test images are also poisoned in the same way for performance evaluation.

Attack settings for other datasets are summarized in Tab.~\ref{tab:attack_config_other_datasets}.
Due to the insufficiency of data, we only conduct all-to-one attacks on these datasets for effective attacks.
We arbitrarily choose class 0 as the target class for CIFAR-100, GTSRB, TinyImageNet, and VGGface2, and class 9 for ImageNette.
All the classes other than the target class are source classes of the attack.
For CIFAR-100, we use the same BadNet, $l_0$ inv, and $l_2$ inv triggers as for CIFAR-10. We increase the perturbation size to 6/255 for the CB pattern for an effective backdoor attack. For each attack, we poison 10 images per source class using the above triggers.
Triggers SP and WaNet are not considered since we cannot launch a successful backdoor attack using these triggers on CIFAR-100.
For GTSRB, in addition to the same triggers as for CIFAR-100, we also use the warping-based trigger (WaNet). We poison 2\% of the training images per source class using the BadNet and $l_2$ inv triggers, and 5\% of the training images per source class with the CB and $l_0$ inv triggers. To achieve similarly effective attacks, we embed the WaNet trigger into 24\% of  the training images per source class.
For TinyImageNet, ImageNette, and VGGFace2, we only consider BadNet as the trigger, as the DNN cannot learn the backdoor mapping using the other (relatively simple and small) triggers in datasets that are much more complicated than CIFAR-10.
To successfully plant backdoors, we increase the size the the BadNet patch to $6\times 6$ for TinyImageNet and to $21\times 21$ for ImageNette and VGGFace2.
We embed the trigger in 10 training images per source class in TinyImageNet and VGGFace2, and in 5\% of the training images per source class for ImageNette.

\subsection{Training Settings}\label{sec:training_settings}
Training settings for the 5 datasets are shown in Table~\ref{tab:training_config}, except for the case of CIFAR-10 poisoned by the CL attack.
We train a ResNet-18 (\cite{ResNet}) on CIFAR-10 and CIFAR-100 for 30 epochs and 40 epochs, respectively.
We train a ResNet-34 (\cite{ResNet}) on both TinyImageNet and ImageNette for 90 epochs.
For GTSRB, we train a MobileNet (\cite{MobileNet}) for 60 epochs.
For VGGFace2, we fine-tune a pre-trained VGG-16 model \citep{VGG} with batch normalization for 90 epochs.
For all models, we use the Adam optimizer (\cite{Adam}) for parameter learning and a scheduler to decay the learning rate of each parameter group by 0.1 every ``scheduler step size'' epochs (shown in the table).
We choose batch size 32 for both CIFAR-10 and CIFAR-100, 64 for GTSRB and ImageNette, and 128 for TinyImageNet.
For CIFAR-10 poisoned by the CL attack, we train a ResNet-18 model following the training settings in the CL paper \cite{CL} for an effective attack. Specifically, we use a stochastic gradient descent (SGD) optimizer for parameter learning, with a momentum of 0.9, a weight decay of 0.0002, and batch size of 50. The initial learning rate of 0.1 is divided by 10 every 30 epochs. The model is trained for 100 epochs.

\begin{table}[h!]
\centering
\scriptsize
\begin{tabular}{cccccccc}
\toprule
\hline
\multirow{2}{*}{\makecell[c]{Trigger type}} &  \multicolumn{2}{c}{BadNet} & \multicolumn{2}{c}{CB} & \multicolumn{2}{c}{$l_0$ inv} & Blend \\
\cline{2-8}
& A2O & A2A & A2O & A2A & A2O & A2A & A2O\\ 
\hline
\makecell[c]{\# poisoned \\ per class}
& 100 & 300 & 100 & 300 & 100 & 300 & 100 \\
$l_0$ norm & $3 \times 3$ & $3 \times 3$ & NA & NA & $1 \times 6$ & $1 \times 6$ & NA \\
$l_2$ norm & NA & NA & 0.3074 & 0.3074 & NA & NA & NA \\
\hline
\multirow{2}{*}{\makecell[c]{Trigger type}} &  \multicolumn{2}{c}{$l_2$ inv} & \multicolumn{2}{c}{SP} & \multicolumn{2}{c}{WaNet} & SIG\\
\cline{2-8}
& A2O & A2A & A2O & A2A & A2O & A2A & A2O\\ 
\hline
\makecell[c]{\# poisoned \\ per class}
& 100 & 300 & 100 & 800 & 900 & 1500 & 100 \\
$l_0$ norm & NA & NA & NA & NA & NA & NA & NA \\
$l_2$ norm & 1.6106 & 1.6106 & 0.5094 & 0.5094 & NA & NA & NA\\
\hline
\bottomrule
\end{tabular}
\caption{Attack configurations on CIFAR-10}
\label{tab:attack_config_CIFAR-10}
\end{table}

\begin{table}[h!]
\centering
\scriptsize
\begin{tabular}{ccccccc}
\toprule
\hline
Dataset & CIFAR-10 & CIFAR-100 & TinyImageNet & ImageNette & GTSRB & VGGFace2 \\
\hline
DNN architecture & ResNet-18 & ResNet-18 & ResNet-34 & ResNet-34 & MobileNet & VGG-16 \\
Optimizer & Adam & Adam & Adam & Adam  & Adam & Adam \\
Batch size & 32 & 32 & 128 & 64 & 64 & 32 \\
Epochs & 30 & 40 & 90 & 90 & 60 & 90 \\
Initial learning rate & 1e-3 & 1e-3 & 1e-3 & 1e-3 & 1e-3 & 1e-3 \\
scheduler step size & 10 & 10 & 30 & 30 & 20 & 30 \\
\hline
\bottomrule
\end{tabular}
\caption{Training configurations of the 6 datasets used in our experiments.}
\label{tab:training_config}
\end{table}

\begin{table}[h!]
\centering
\scriptsize
\begin{tabular}{ccccccc}
\toprule
\hline
\multirow{2}{*}{\makecell[c]{Trigger type}} & \multicolumn{4}{c}{CIFAR-100} & TinyImageNet & ImageNette \\
\cline{2-7}
& BadNet & CB & $l_0$ inv & $l_2$ inv & BadNet & BadNet \\
\hline
Target class & 0 & 0 & 0 & 0 & 0 & 9 \\
\makecell[c]{\# poisoned \\ per class}
& 10 & 10 & 10 & 10 & 10 & 5\% \\
$l_0$ norm
& $3 \times 3$ & NA & $1 \times 6$ & NA & $6 \times 6$ & $21 \times 21$ \\
$l_2$ norm
& NA & 0.9222 & NA & 1.6106 & NA & NA \\
\hline
\multirow{2}{*}{\makecell[c]{Trigger type}} & \multicolumn{5}{c}{GTSRB} & VGGFace2 \\
\cline{2-7}
& BadNet & CB & $l_0$ inv & $l_2$ inv & WaNet & BadNet \\
\hline
Target class & 0 & 0 & 0 & 0 & 0 & 0 \\
\makecell[c]{\# poisoned \\ per class}
& 2\% & 5\% & 5\% & 2\% & 24\% & 10 \\
$l_0$ norm
& $3 \times 3$ & NA & $1 \times 6$ & NA & NA & $21 \times 21$ \\
$l_2$ norm
& NA & 0.9222 & NA & 1.6106 & NA & NA \\
\hline
\bottomrule
\end{tabular}
\caption{Attack configurations on GTSRB, CIFAR-100, ImageNette, TinyImageNet, and VGGFace2.}
\label{tab:attack_config_other_datasets}
\end{table}

\section{Pattern Estimation and Backdoor Detection}\label{sec:backdoor_detection}

For BNA, following Sec. \ref{subsec:backdoor_mitigation_method}, we first perform detection by reverse-engineering a backdoor trigger for each class pair.
For patch triggers like BadNet, we use the objective function from \cite{NC} for trigger reverse-engineering. 
For other more subtle, perturbation-based trigger types, we use the objective function from \cite{TNNLS} for reverse-engineering.
The detection statistic is the reciprocal of the $l_0$ norm of the relaxed masks of the estimated patch triggers and $l_2$ norm of reverse-engineered perturbation-based triggers.
Then we feed the statistics obtained from the estimated trigger to an anomaly detector. 

\begin{figure}[!ht]
	\centering
	\begin{subfigure}[]{.49\textwidth}
		\centering
		\includegraphics[width=\textwidth]{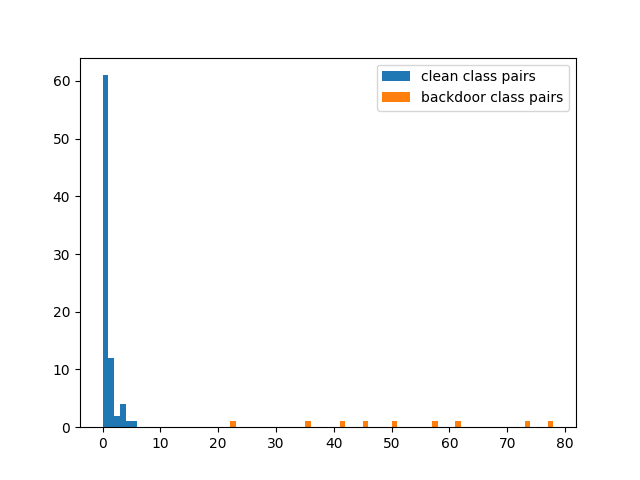}
		 \caption{BadNet}
	\end{subfigure}
    \hfill
    \begin{subfigure}[]{.49\textwidth}
         \centering
         \includegraphics[width=\textwidth]{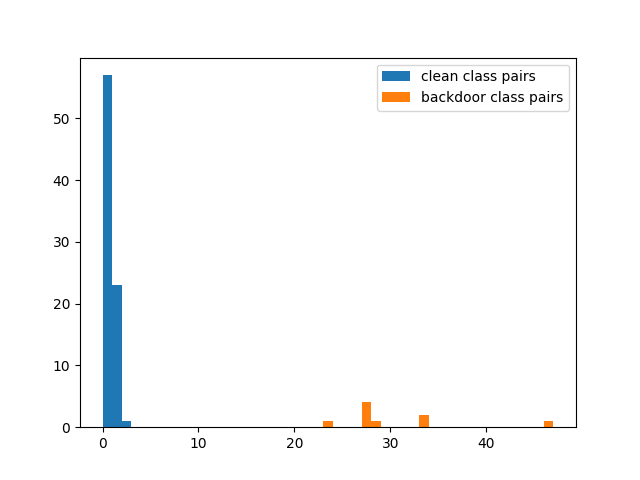}
         \caption{CB}
     \end{subfigure}
     \hfill
    \begin{subfigure}[]{.49\textwidth}
         \centering
         \includegraphics[width=\textwidth]{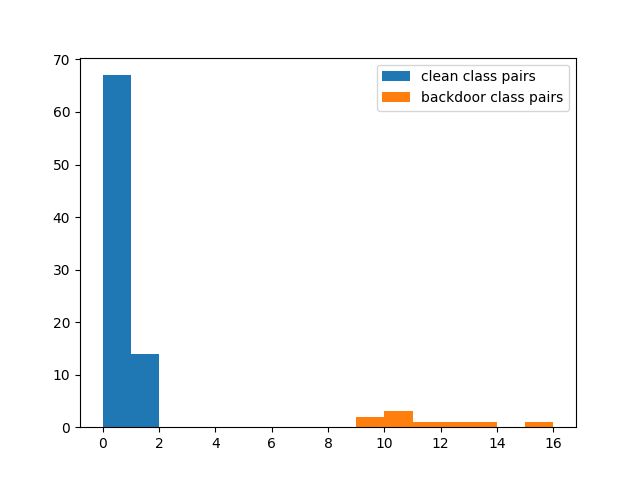}
         \caption{SP}
     \end{subfigure}
     \hfill
    \begin{subfigure}[]{.49\textwidth}
         \centering
         \includegraphics[width=\textwidth]{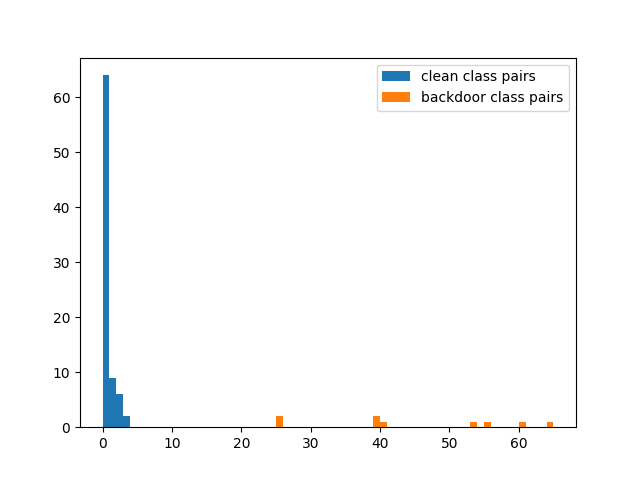}
         \caption{$l_0$ inv}
     \end{subfigure}
     \hfill
    \begin{subfigure}[]{.49\textwidth}
         \centering
         \includegraphics[width=\textwidth]{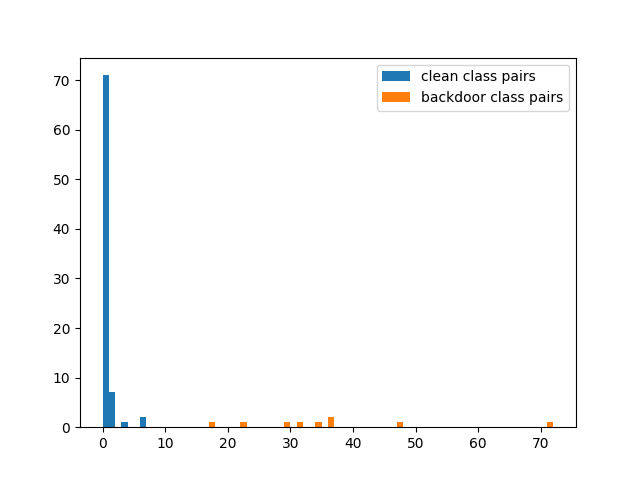}
         \caption{$l_2$ inv}
     \end{subfigure}
     \hfill
    \begin{subfigure}[]{.49\textwidth}
         \centering
         \includegraphics[width=\textwidth]{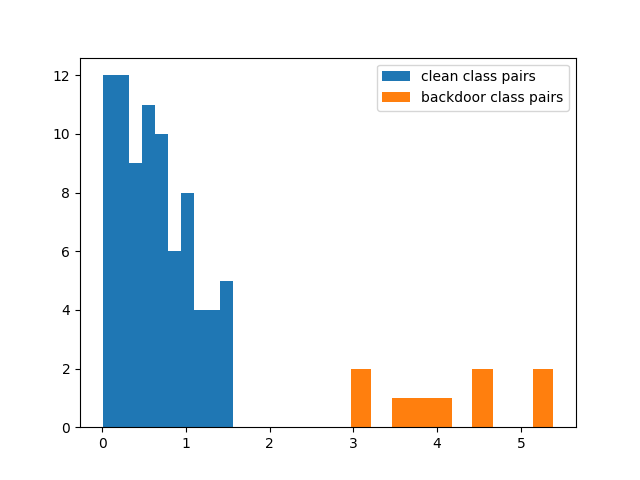}
         \caption{WaNet}
         \label{fig:detect_hist_WaNet_all2one}
     \end{subfigure}
	\caption{Histograms of anomaly scores for each class pair under all all-to-one attacks.}
	\label{fig:outlier_detection_all2one}
\end{figure}

\begin{figure}[h!]
	\centering
    \begin{subfigure}[]{.49\textwidth}
         \centering
         \includegraphics[width=\textwidth]{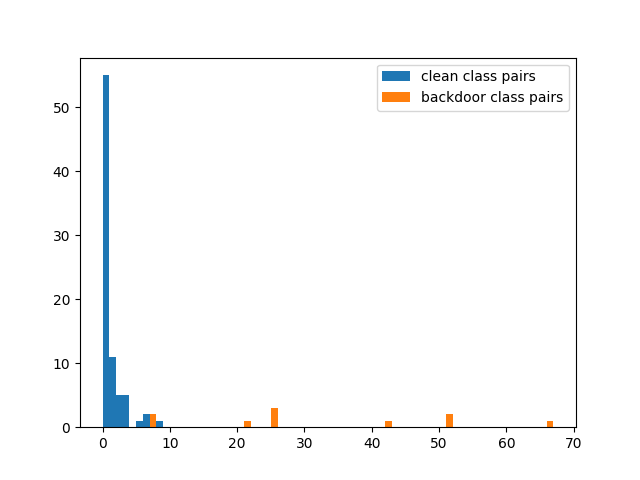}
         \caption{BadNet}
         \label{fig:detect_hist_BadNet_all2all}
    \end{subfigure}
    \hfill
    \begin{subfigure}[]{.49\textwidth}
         \centering
         \includegraphics[width=\textwidth]{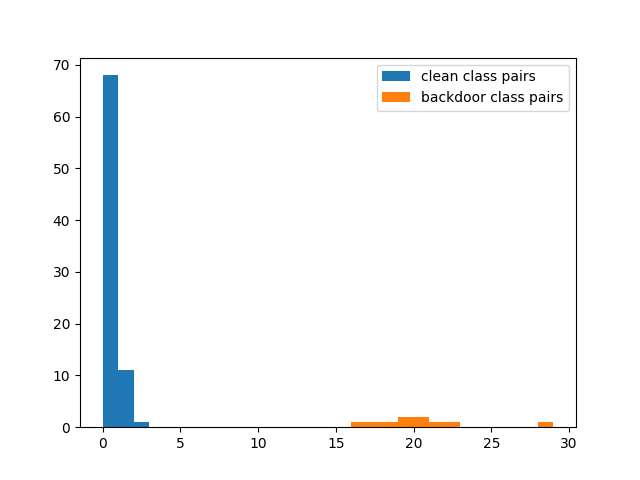}
         \caption{CB}
     \end{subfigure}
     \hfill
    \begin{subfigure}[]{.49\textwidth}
         \centering
         \includegraphics[width=\textwidth]{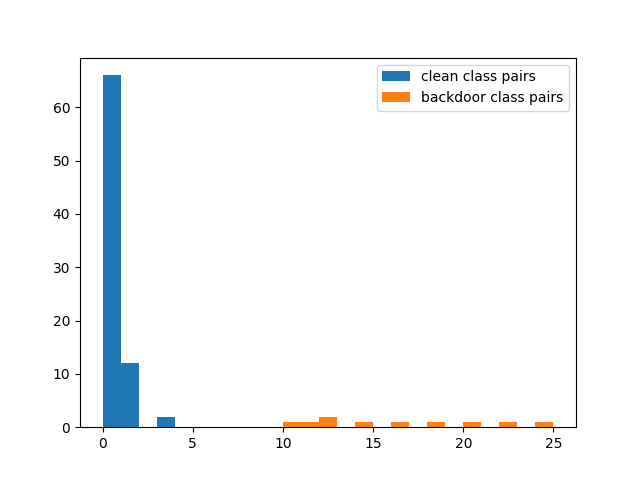}
         \caption{SP}
     \end{subfigure}
     \hfill
    \begin{subfigure}[]{.49\textwidth}
         \centering
         \includegraphics[width=\textwidth]{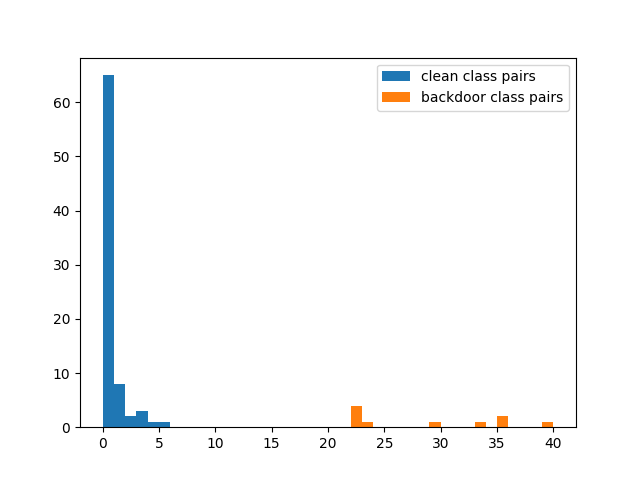}
         \caption{$l_0$ inv}
     \end{subfigure}
     \hfill
    \begin{subfigure}[]{.49\textwidth}
         \centering
         \includegraphics[width=\textwidth]{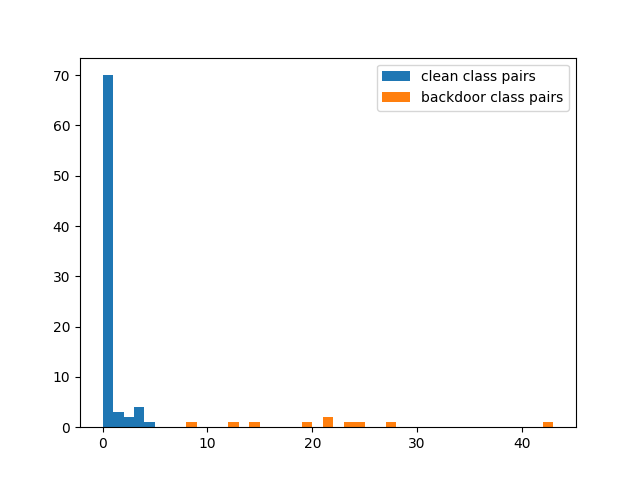}
         \caption{$l_2$ inv}
     \end{subfigure}
     \hfill
    \begin{subfigure}[]{.49\textwidth}
         \centering
         \includegraphics[width=\textwidth]{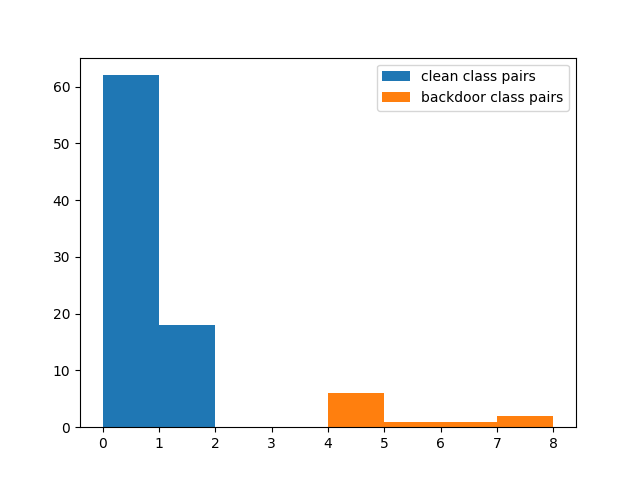}
         \caption{WaNet}
         \label{fig:detect_hist_WaNet_all2all}
     \end{subfigure}
	\caption{Histograms of anomaly scores for each class pair under all all-to-all attacks.}
	\label{fig:outlier_detection_all2all}
\end{figure}

Our anomaly detector is based on MAD, which is a classical approach also used by \cite{NC, DeepInspect, DataLimited}. 
It first calculates the absolute deviation between all detection statistics (the reciprocal of $l_0$ norm of patch triggers and $l_2$ norm of perturbation-based triggers) and the median, and the median of the absolute deviations is called Median Absolute Deviation (MAD).
For a class pair and its corresponding estimated trigger, if the trigger's anomaly score, which is defined as the absolute deviation divided by MAD, is larger than a given threshold, it is detected as a backdoor class pair. 
A wide range of 
detection thresholds detect the attack, as shown in Fig.~\ref{fig:outlier_detection_all2one} and Fig.~\ref{fig:outlier_detection_all2all}. 
These figures show the histograms of the anomaly scores for all class pairs under all-to-one and all-to-all attacks, respectively.
Here, we set the detection threshold at 7, which easily catches all the backdoor class pairs under all the attacks, except for the all-to-all BadNet attack and both attacks using the WaNet trigger.

For the all-to-all BadNet attack, the outlier detector finds two source classes -- 0 and 8 -- for the  target class 1, where 0-1 is the true source-target class pair and 8-1 is falsely detected, as shown in Fig.~\ref{fig:detect_hist_BadNet_all2all}.
The $l_0$ norm of the trigger estimated on class 0 clean images is 3.02, and that estimated on class 8 images is 7.95.
If class 0 and 8 are both the source classes involved in the backdoor attack, then the trigger estimated on the clean images from class 0 \textit{and} 8 should both have a small $l_0$ norm. Otherwise, the trigger estimated using class 8 images is an intrinsic backdoor pattern (\cite{XiangMCLK22, Liu_2022_CVPR, Tao_2022_CVPR}), and 0-1 is the true source class pair, since the trigger of 0-1 has smaller size than 8-1.
By optimizing on clean images from class 0 \textit{and} 8, the $l_0$ norm of the trigger that causes mis-classification to class 1 with high confidence is 27.18 -- much larger than the triggers estimated on either class 0 images or class 8 images.
Thus, we detect 0-1 as the true backdoor class pair and discard the trigger for class pair 8-1 in backdoor mitigation.

For the attacks using warping-based triggers (WaNet), unlike the other attacks, the trigger size for clean class pairs and backdoor class pairs are both small. However, there is still a ``gap'' between the anomaly scores of clean class pairs and backdoor class pairs, as shown in Fig.~\ref{fig:detect_hist_WaNet_all2one} and \ref{fig:detect_hist_WaNet_all2all}. 
The outlier detector successfully detects all the backdoor class pairs by using a threshold at 3.


\section{Distribution divergences}\label{sec:distribution_div}
As stated in Thm.~\ref{thm:main}, for our backdoor mitigation method, SIA monotonically increases as the divergence between clean instances and backdoor-trigger instances decreases.
We show the ACC, ASR, and SIA for our method against all all-to-one attacks on CIFAR-10 in Tab.~\ref{tab:comparison}.
We also show the corresponding distribution divergences under all attacks in Tab.~\ref{tab:distribution_divergences}.
Tab.~\ref{tab:distribution_divergences} shows the average TV distance, JS divergence, and KL divergence between distributions of penultimate layer activations of clean images and backdoor-trigger images in clean ResNet-18 (Clean), backdoor poisoned ResNet-18 (Poisoned), backdoor poisoned ResNet-18 mitigated by NC (NC), and backdoor poisoned ResNet-18 mitigated by BNA (BNA).
For the backdoor poisoned ResNet-18 mitigated by BNA, we use TV divergence in backdoor mitigation.
For all attacks, all three divergences are small for a clean DNN, while relatively large for a backdoor-poisoned DNN. The distribution of backdoor-trigger instances severely deviates from that of clean instances.
With our mitigation method (BNA), the distribution alteration is significantly relieved. All the three divergences are drastically reduced, which is consistent with the results in Tab.~\ref{tab:comparison}.
NC also relieves distribution alteration -- it even performs better than ours with regard to distribution divergences.
However, NC cannot achieve SIA as high as our BNA (see Tab.~\ref{tab:comparison}).
It tunes model parameters based on \textit{insufficient} data (\ie, the data assumed to be available to the defender); therefore the ACC of the DNN drops after parameter tuning. 
The achievable SIA of NC also drops since it is upper-bouned by the ACC.

\begin{table}[h!]
\centering
\scriptsize
\begin{tabular}{cllllll}
\toprule
\hline
Trigger type & BadNet & CB & $l_0$ inv & $l_2$ inv & SP & WaNet \\
\hline
\multicolumn{7}{c}{KL divergence					} \\
\hline
Clean & 0.0022 & 0.0010 & 0.0013 & 0.0085 & 0.0017 & 0.0037 \\ 
Poisoned & 0.3211 & 0.752 & 0.4029 & 0.8730 & 0.4385 & 0.2708 \\ 
BNA & 0.0291 & 0.0141 & 0.0337 & 0.0402 & 0.0146 & 0.0389 \\
NC & 0.0045 & 0.0044 & 0.0092 & 0.0841 & 0.0174 & 0.0077  \\
\hline
\multicolumn{7}{c}{JS divergence					} \\
\hline
Clean & 0.0239 & 0.0166 & 0.0185 & 0.0461 & 0.0214 & 0.0311 \\ 
Poisoned & 0.2867 & 0.3528 & 0.3275 & 0.3898 & 0.3041 & 0.2215 \\ 
BNA & 0.0952 & 0.0583 & 0.0847 & 0.0986 & 0.0582 & 0.0611 \\
NC & 0.0362 & 0.0350 & 0.0477 & 0.1440 & 0.0604 & 0.0450 \\
\hline
\multicolumn{7}{c}{TV distance					} \\
\hline
Clean & 634.1953 & 376.4219 & 461.9531 & 1240.3359 & 554.3594 & 805.9648 \\ 
Poisoned & 8877.2734 & 11177.3867 & 9203.9609 & 12482.4336 & 10119.3281 & 7237.7812 \\
BNA & 2463.582 & 1705.0156 & 2490.457 & 2794.7578 & 1761.2148 & 1749.2461 \\
NC & 774.78 & 822.13 & 1284.78 & 3799.22 & 1905.65 & 1255.89 \\
\hline
\bottomrule
\end{tabular}
\caption{Average TV distance, JS divergence, and KL divergence between distributions of clean instances and backdoor-trigger instances in clean DNN, poisoned DNN, and poisoned DNN mitigated by BNA using TV divergence.}
\label{tab:distribution_divergences}
\end{table}

\section{Choice of divergence forms}\label{sec:choice_loss_func}
In Tab.~\ref{tab:comparison} and \ref{tab:other datasets}, we only show the results for BNA using TV distance in backdoor mitigation (Eq.~\ref{eq:opt_raw}).
Here we show that our method is not sensitive to the choice of distribution divergence form.
We respectively use TV distance, JS divergence, and KL divergence to mitigate the 5 all-to-one CB attacks against CIFAR-10, and show the average distribution similarity measured by the three measurements after mitigation. 
The distribution similarity is calculated on the penultimate layer activations.
As shown in Tab.~\ref{tab:choice_loss_func}, the distribution alteration is significantly relieved after mitigation, regardless of the divergence form used in mitigation (Eq.~\ref{eq:opt_raw}).

\begin{table}[h!]
\centering
\scriptsize
\begin{tabular}{c|lll}
\toprule
\hline
{\makecell[c]{divergence form used in mitigation $\rightarrow$ \\ distribution similarity after mitigation $\downarrow$}}  & TV & JS & KL \\
\hline
TV & 1742 & 1730 & 1719 \\
JS & 0.0622 & 0.06172 & 0.0613 \\
KL & 0.0165 & 0.0163 & 0.0161 \\
\hline
\bottomrule
\end{tabular}
\caption{Average TV, JS, and KL between clean instances and backdoor-trigger instances using TV, JS, and KL for measuring distribution similarity in BNA-based backdoor mitigation.}
\label{tab:choice_loss_func}
\end{table}

\section{Impact of perturbation size and poisoning ratio on backdoor mitigation}\label{sec:bd_poison_ratio_pert_size}

To observe the impact of attack settings on the performance of backdoor mitigation methods, we tune the poisoning ratio (\ie, the number of poisoned instances per source class) and perturbation size used in all-to-one CB attacks, and apply all the mitigation methods on these poisoned DNNs.
The results are shown in Tab.~\ref{tab:bd_poison_ratio_pert_size_RW}.
Generally, the metrics for all methods decrease with increasing poisoning ratio and perturbation size.
Although the performance for our BNA slightly declines as the attack is strengthened, our method still outperforms other methods in terms of SIA. 
BNA also achieves the best or comparable ACC and ASR to other methods.

\begin{table}[h!]
\centering
\scriptsize
\begin{tabular}{cc|ccccc|ccccc}
\toprule
\hline
\multicolumn{2}{c|}{Mitigation method} & \multicolumn{5}{c|}{the number of poisoned instances per class} & \multicolumn{5}{c}{perturbation size (*255)} \\
& & 50 & 100 & 150 & 200 & 250 & 2 & 3 & 4 & 5 & 6 \\
\hline  
\multirow{3}{*}{NC}
& ACC & 0.8953 & 0.8734 & 0.8826 & 0.8943 & 0.8799 & 0.8734 & 0.8918 & 0.8667 & 0.8825 & 0.8709 \\ 
& ASR & \textbf{0.0056} & 0.0238 & 0.0148 & \textbf{0.0057} & 0.0064 & 0.0238 & \textbf{0.0042} & 0.0157 & 0.0133 & \textbf{0.0065}\\ 
& SIA & 0.8515 & 0.8412 & 0.8552 & 0.8579 & 0.8532 & 0.8412 & 0.8560 & 0.8283 & 0.8194 & 0.7883 \\
\hline  
\multirow{3}{*}{I-BAU}
& ACC & 0.8708 & 0.8473 & 0.8818 & 0.8941 & 0.8594 & 0.8473 & 0.8646 & 0.8822 & \textbf{0.9004} & 0.8919 \\ 
& ASR & 0.0789 & \textbf{0.0043} & 0.0712 & 0.5074 & 0.0460 & \textbf{0.0043} & 0.3052 & 0.0247 & \textbf{0.0011} & 0.2048 \\ 
& SIA & 0.6564 & 0.8399 & 0.7563 & 0.3802 & 0.6715 & 0.8399 & 0.5823 & 0.7712 & 0.8102 & 0.5153 \\
\hline  
\multirow{3}{*}{ANP}
& ACC & 0.8523 & 0.8271 & 0.8612 & 0.8204 & 0.7614 & 0.8271 & 0.8486 & 0.8156 & 0.8418 & 0.8249 \\ 
& ASR & 0.1940 & 0.8535 & 0.5401 & 0.0031 & \textbf{0.0043} & 0.8535 & 0.9836 & 0.6394 & 0.3670 & 0.2548 \\ 
& SIA & 0.5158 & 0.1047 & 0.2440 & 0.6157 & 0.3701 & 0.1047 & 0.0142 & 0.1911 & 0.3238 & 0.4606 \\
\hline  
\multirow{3}{*}{NAD}
& ACC & 0.8942 & 0.8745 & 0.8949 & 0.8902 & 0.8674 & 0.8767 & 0.8823 & 0.8745 & 0.8835 & \textbf{0.8990} \\ 
& ASR & 0.0147 & 0.0086 & \textbf{0.0095} & 0.0106 & 0.0125 & 0.0070 & 0.0096 & 0.0086 & 0.0642 & 0.0586 \\ 
& SIA & 0.8646 & 0.8504 & 0.8709 & 0.8695 & 0.8514 & 0.8631 & 0.8574 & 0.8504 & 0.8070 & 0.7896 \\
\hline  
\multirow{3}{*}{ARGD}
& ACC & 0.8743 & 0.8832 & 0.8693 & 0.8659 & 0.8415 & 0.8832 & 0.8872 & 0.8619 & 0.8508 & 0.8394 \\ 
& ASR & 0.0106 & 0.0108 & 0.0097 & 0.0117 & 0.0121 & 0.0108 & 0.0073 & \textbf{0.0083} & 0.0085 & 0.0153 \\ 
& SIA & 0.8590 & 0.8685 & 0.8528 & 0.8482 & 0.8267 & 0.8685 & 0.8574 & 0.8467 & 0.8373 & 0.8196 \\
\hline  
\multirow{3}{*}{\makecell[c]{BNA \\ (ours)}}
& ACC & \textbf{0.9112} & \textbf{0.9094} & \textbf{0.9098} & \textbf{0.9102} & \textbf{0.9015} & \textbf{0.9094} & \textbf{0.9041} & \textbf{0.9079} & 0.8992 & 0.8912 \\ 
& ASR & 0.0095 & 0.0141 & 0.0121 & 0.0170 & 0.0090 & 0.0141 & 0.0395 & 0.0222 & 0.0109 & 0.0388 \\ 
& SIA & \textbf{0.8851} & \textbf{0.8837} & \textbf{0.8728} & \textbf{0.8840} & \textbf{0.8662} & \textbf{0.8851} & \textbf{0.8783} & \textbf{0.8711} & \textbf{0.8814} & \textbf{0.8435} \\
\hline
\bottomrule
\end{tabular}
\caption{ACC, ASR, and SIA for BNA, NC, I-BAU, ANP, NAD, and ARGD as a function of (1) the number of poisoned instances injected into the training set; (2) the perturbation size under all-to-one CB attack.}
\label{tab:bd_poison_ratio_pert_size_RW}
\end{table}



\clearpage
\bibliographystyle{APA}

\begin{thebibliography}{100}
\providecommand{\natexlab}[1]{#1}
\expandafter\ifx\csname urlstyle\endcsname\relax
  \providecommand{\doi}[1]{doi:\discretionary{}{}{}#1}\else
  \providecommand{\doi}{doi:\discretionary{}{}{}\begingroup
  \urlstyle{rm}\Url}\fi

\bibitem[{Bengio, Y. et~al.(2007) Bengio, Y. \& LeCun, Y.}]{Bengio+chapter2007}
Bengio, Y. \& LeCun, Y. (2007).
\newblock Scaling Learning Algorithms Towards AI. 
\newblock {\em Large Scale Kernel Machines}. 

\bibitem[{Hinton, G. et~al.(2006) Hinton, G., Osindero, S. \& Teh, Y.}]{Hinton06}
Hinton, G., Osindero, S. \& Teh, Y. (2006).
\newblock A Fast Learning Algorithm for Deep Belief Nets. 
\newblock {\em Neural Computation}. \textbf{18} pp. 1527-1554 

\bibitem[{Goodfellow, I. et~al.(2016) Goodfellow, I., Bengio, Y., Courville, A. \& Bengio, Y.}]{goodfellow2016deep}
Goodfellow, I., Bengio, Y., Courville, A. \& Bengio, Y. (2016).
\newblock Deep learning. 
\newblock {\em MIT Press}

\bibitem[{Li, Y. et~al.(2022) Li, Y., Jiang, Y., Li, Z. \& Xia, S.}]{BAsurvey}
Li, Y., Jiang, Y., Li, Z. \& Xia, S. (2022).
\newblock Backdoor Learning: A Survey. 
\newblock {\em IEEE Transactions On Neural Networks And Learning Systems}. pp. 1-18 

\bibitem[{Ilyas, A. et~al.(2019) Ilyas, A., Santurkar, S., Tsipras, D., Engstrom, L., Tran, B. \& Madry, A.}]{Ilyas_NIPS_2019}
Ilyas, A., Santurkar, S., Tsipras, D., Engstrom, L., Tran, B. \& Madry, A. (2019).
\newblock Adversarial Examples Are Not Bugs, They Are Features. 
\newblock {\em NeurIPS}. 

\bibitem[{Frankle, J. et~al.(2021) Frankle, J., Schwab, D. \& Morcos, A.}]{FrankleSM21}
Frankle, J., Schwab, D. \& Morcos, A. (2021).
\newblock Training BatchNorm and Only BatchNorm: On the Expressive Power of Random Features in CNNs. 
\newblock {\em ICLR}. 

\bibitem[{Li, X. et~al.(2022) Li, X., Xiang, Z., Miller, D. \& Kesidis, G.}]{InFlight}
Li, X., Xiang, Z., Miller, D. \& Kesidis, G. (2022).
\newblock Test-Time Detection of Backdoor Triggers for Poisoned Deep Neural Networks. 
\newblock {\em ICASSP}. 

\bibitem[{Benz, P. et~al.(2021) Benz, P., Zhang, C., Karjauv, A. \& Kweon, I.}]{Benz_2021_WACV}
Benz, P., Zhang, C., Karjauv, A. \& Kweon, I. (2021).
\newblock Revisiting Batch Normalization for Improving Corruption Robustness. 
\newblock {\em WACV}. 

\bibitem[{Zhao, Z. et~al.(2022) Zhao, Z., Chen, X., Xuan, Y., Dong, Y., Wang, D. \& Liang, K. }]{Zhao_2022_CVPR}
Zhao, Z., Chen, X., Xuan, Y., Dong, Y., Wang, D. \& Liang, K. (2022).
\newblock DEFEAT: Deep Hidden Feature Backdoor Attacks by Imperceptible Perturbation and Latent Representation Constraints. 
\newblock {\em CVPR}. 

\bibitem[{Qi, X. et~al.(2022) Qi, X., Xie, T., Pan, R., Zhu, J., Yang, Y. \& Bu, K.}]{Qi_2022_CVPR}
Qi, X., Xie, T., Pan, R., Zhu, J., Yang, Y. \& Bu, K. (2022).
\newblock Towards Practical Deployment-Stage Backdoor Attack on Deep Neural Networks. 
\newblock {\em CVPR}. 

\bibitem[{Xie, C.et~al.(2020) Xie, C., Huang, K., Chen, P. \& Li, B.}]{Xie2020DBA}
Xie, C., Huang, K., Chen, P. \& Li, B. (2020).
\newblock DBA: Distributed Backdoor Attacks against Federated Learning. 
\newblock {\em ICLR}. 

\bibitem[{Wang, Z. et~al.(2022) Wang, Z., Zhai, J. \& Ma, S.}]{Wang_2022_CVPR}
Wang, Z., Zhai, J. \& Ma, S. (2022).
\newblock BppAttack: Stealthy and Efficient Trojan Attacks Against Deep Neural Networks via Image Quantization and Contrastive Adversarial Learning. 
\newblock {\em CVPR}. 

\bibitem[{Yao, Y. et~al.(2019) Yao, Y., Li, H., Zheng, H. \& Zhao, B.}]{Yao_2019_CCS}
Yao, Y., Li, H., Zheng, H. \& Zhao, B. (2019).
\newblock Latent Backdoor Attacks on Deep Neural Networks. 
\newblock {\em Proceedings Of The 2019 ACM SIGSAC Conference On Computer And Communications Security}. 

\bibitem[{Wang, L. et~al.(2021) Wang, L., Javed, Z., Wu, X., Guo, W., Xing, X. \& Song, D.}]{Wang_2021_IJCAI}
Wang, L., Javed, Z., Wu, X., Guo, W., Xing, X. \& Song, D. (2021).
\newblock BACKDOORL: Backdoor Attack against Competitive Reinforcement Learning. 
\newblock {\em IJCAI}. 

\bibitem[{Xiang, Z. et~al.(2019) Xiang, Z., Miller, D. \& Kesidis, G.}]{CI}
Xiang, Z., Miller, D. \& Kesidis, G. (2019).
\newblock A benchmark study of backdoor data poisoning defenses for deep neural network classifiers and a novel defense. 
\newblock {\em MLSP}. 

\bibitem[{Zhang, Z. et~al.(2022) Zhang, Z., Lyu, L., Wang, W., Sun, L. \& Sun, X.}]{zhang2022how}
Zhang, Z., Lyu, L., Wang, W., Sun, L. \& Sun, X. (2022).
\newblock How to Inject Backdoors with Better Consistency: Logit Anchoring on Clean Data. 
\newblock {\em ICLR}. 

\bibitem[{Schneider, S. et~al.(2020) Schneider, S., Rusak, E., Eck, L., Bringmann, O., Brendel, W. \& Bethge, M.}]{SchneiderRE0BB20}
Schneider, S., Rusak, E., Eck, L., Bringmann, O., Brendel, W. \& Bethge, M. (2020).
\newblock Improving robustness against common corruptions by covariate shift adaptation. 
\newblock {\em NeurIPS}. 

\bibitem[{Santurkar, S.et~al.(2018) Santurkar, S., Tsipras, D., Ilyas, A. \& Madry, A.}]{SanturkarTIM18}
Santurkar, S., Tsipras, D., Ilyas, A. \& Madry, A. (2018).
\newblock How Does Batch Normalization Help Optimization?. 
\newblock {\em NeurIPS}. 

\bibitem[{Lubana, E.et~al.(2021) Lubana, E., Dick, R. \& Tanaka, H.}]{lubana2021beyond}
Lubana, E., Dick, R. \& Tanaka, H. (2021).
\newblock Beyond BatchNorm: Towards a Unified Understanding of Normalization in Deep Learning. 
\newblock {\em NeurIPS}. 

\bibitem[{Ioffe, S. et~al.(2015) Ioffe, S. \& Szegedy, C.}]{IoffeS15}
Ioffe, S. \& Szegedy, C. (2015).
\newblock Batch Normalization: Accelerating Deep Network Training by Reducing Internal Covariate Shift. 
\newblock {\em ICML}. 

\bibitem[{Li, X. et~al.(2021) Li, X., Kesidis, G., Miller, D. \& Lucic, V.}]{LKML21}
Li, X., Kesidis, G., Miller, D. \& Lucic, V. (2021).
\newblock Backdoor Attack and Defense for Deep Regression. 
\newblock {\em ArXiv}. 2109.02381

\bibitem[{Krizhevsky, A.(2009)}]{cifar10}
Krizhevsky, A. (2009).
\newblock Learning multiple layers of features from tiny images. 
\newblock \url{http://www.cs.toronto.edu/ kriz/learning-features-2009-TR.pdf}

\bibitem[{Wang, B. et~al.(2019) Wang, B., Yao, Y., Shan, S., Li, H., Viswanath, B., Zheng, H. \& Zhao, B.}]{NC}
Wang, B., Yao, Y., Shan, S., Li, H., Viswanath, B., Zheng, H. \& Zhao, B. (2019).
\newblock Neural Cleanse: Identifying and Mitigating Backdoor Attacks in Neural Networks. 
\newblock {\em 2019 IEEE Symposium On Security And Privacy}. 

\bibitem[{Gao, Y. et~al.(2019) Gao, Y., Xu, C., Wang, D., Chen, S., Ranasinghe, D. \& Nepal, S.}]{STRIP}
Gao, Y., Xu, C., Wang, D., Chen, S., Ranasinghe, D. \& Nepal, S. (2019).
\newblock STRIP: a defence against trojan attacks on deep neural networks. 
\newblock {\em ACSAC}. 

\bibitem[{Liu, Y. et~al.(2018) Liu, Y., Ma, S., Aafer, Y., Lee, W., Zhai, J., Wang, W. \& Zhang, X.}]{Trojan}
Liu, Y., Ma, S., Aafer, Y., Lee, W., Zhai, J., Wang, W. \& Zhang, X. (2018).
\newblock Trojaning Attack on Neural Networks. 
\newblock {\em NDSS}. 

\bibitem[{Gu, T. et~al.(2019) Gu, T., Liu, K., Dolan-Gavitt, B. \& Garg, S.}]{BadNets}
Gu, T., Liu, K., Dolan-Gavitt, B. \& Garg, S. (2019).
\newblock BadNets: Evaluating Backdooring Attacks on Deep Neural Networks. 
\newblock {\em IEEE Access}. 

\bibitem[{Turner, A. et~al.(2019) Turner, A., Tsipras, D. \& Madry, A.}]{CL}
Turner, A., Tsipras, D. \& Madry, A. (2019).
\newblock Label-Consistent Backdoor Attacks. 
\newblock {\em ArXiv}. 1912.02771

\bibitem[{Chen, X. et~al.(2017) Chen, X., Liu, C., Li, B., Lu, K. \& Song, D.}]{Targeted-Backdoor}
Chen, X., Liu, C., Li, B., Lu, K. \& Song, D. (2017).
\newblock Targeted Backdoor Attacks on Deep Learning Systems Using Data Poisoning. 
\newblock {\em ArXiv}. 1712.05526

\bibitem[{Dong, Y. et~al.(2021) Dong, Y., Yang, X., Deng, Z., Pang, T., Xiao, Z., Su, H. \& Zhu, J.}]{B3D}
Dong, Y., Yang, X., Deng, Z., Pang, T., Xiao, Z., Su, H. \& Zhu, J. (2021).
\newblock Black-box Detection of Backdoor Attacks with Limited Information and Data. 
\newblock {\em ICCV}. 

\bibitem[{Doan, K.et~al.(2021) Doan, K., Lao, Y. \& Li, P.}]{WB}
Doan, K., Lao, Y. \& Li, P. (2021).
\newblock Backdoor Attack with Imperceptible Input and Latent Modification. 
\newblock {\em NeurIPS}. 

\bibitem[{Hampel, F.(1974)}]{MAD}
Hampel, F. (1974).
\newblock The influence curve and its role in robust estimation. 
\newblock {\em Journal Of The American Statistical Association}. 

\bibitem[{Chou, E. et~al.(2020) Chou, E., Tramèr, F. \& Pellegrino, G.}]{SentiNet}
Chou, E., Tramèr, F. \& Pellegrino, G. (2020).
\newblock SentiNet: Detecting Localized Universal Attacks Against Deep Learning Systems. 
\newblock {\em 2020 IEEE Security And Privacy Workshops}. 

\bibitem[{Xiang, Z. et~al.(2022) Xiang, Z., Miller, D. \& Kesidis, G.}]{TNNLS}
Xiang, Z., Miller, D. \& Kesidis, G. (2022).
\newblock Detection of Backdoors in Trained Classifiers Without Access to the Training Set. 
\newblock {\em IEEE Transactions On Neural Networks And Learning Systems}. 

\bibitem[{He, K. et~al.(2016) He, K., Zhang, X., Ren, S. \& Sun, J.}]{ResNet}
He, K., Zhang, X., Ren, S. \& Sun, J. (2016).
\newblock Deep Residual Learning for Image Recognition. 
\newblock {\em CVPR}. 

\bibitem[{Howard, A. et~al.(2017) Howard, A., Zhu, M., Chen, B., Kalenichenko, D., Wang, W., Weyand, T., Andreetto, M. \& Adam, H.}]{MobileNet}
Howard, A., Zhu, M., Chen, B., Kalenichenko, D., Wang, W., Weyand, T., Andreetto, M. \& Adam, H. (2017).
\newblock MobileNets: Efficient Convolutional Neural Networks for Mobile Vision Applications. 
\newblock {\em ArXiv}. 1704.04861

\bibitem[{Kingma, D. et~al.(2015) Kingma, D. \& Ba, J.}]{Adam}
Kingma, D. \& Ba, J. (2015).
\newblock Adam: A Method for Stochastic Optimization. 
\newblock {\em ICLR}. 

\bibitem[{Wang, S. et~al.(2020) Wang, S., Nepal, S., Rudolph, C., Grobler, M., Chen, S. \& Chen, T.}]{BackdoorTransferLearning}
Wang, S., Nepal, S., Rudolph, C., Grobler, M., Chen, S. \& Chen, T. (2020).
\newblock Backdoor Attacks against Transfer Learning with Pre-trained Deep Learning Models. 
\newblock {\em IEEE Transactions On Services Computing}. 

\bibitem[{Yao, Y. et~al.(2019) Yao, Y., Li, H., Zheng, H. \& Zhao, B.}]{DBLP:conf/ccs/YaoLZZ19}
Yao, Y., Li, H., Zheng, H. \& Zhao, B. (2019).
\newblock Latent Backdoor Attacks on Deep Neural Networks. 
\newblock {\em Proceedings Of The 2019 ACM SIGSAC Conference On Computer And Communications Security}. 

\bibitem[{Li, S. et~al.(2019) Li, S., Zhao, B., Yu, J., Xue, M., Kaafar, D. \& Zhu, H.}]{DBLP:journals/corr/abs-1909-02742}
Li, S., Zhao, B., Yu, J., Xue, M., Kaafar, D. \& Zhu, H. (2019).
\newblock Invisible Backdoor Attacks Against Deep Neural Networks. 
\newblock {\em ArXiv}. 1909.02742

\bibitem[{Saha, A. et~al.(2020) Saha, A., Subramanya, A. \& Pirsiavash, H.}]{HiddenTrigger}
Saha, A., Subramanya, A. \& Pirsiavash, H. (2020).
\newblock Hidden Trigger Backdoor Attacks. 
\newblock {\em AAAI}. 

\bibitem[{Miller, D. et~al.(2019) Miller, D., Wang, Y. \& Kesidis, G.}]{TTE}
Miller, D., Wang, Y. \& Kesidis, G. (2019).
\newblock When Not to Classify: Anomaly Detection of Attacks (ADA) on DNN Classifiers at Test Time. 
\newblock {\em Neural Comput.}. 

\bibitem[{Chen, X. et~al.(2020) Chen, X., Salem, A., Backes, M., Ma, S. \& Zhang, Y.}]{DBLP:journals/corr/abs-2006-01043}
Chen, X., Salem, A., Backes, M., Ma, S. \& Zhang, Y. (2020).
\newblock BadNL: Backdoor Attacks Against NLP Models. 
\newblock {\em ArXiv}. 2006.01043 

\bibitem[{Dai, J. et~al.(2019) Dai, J., Chen, C. \& Li, Y.}]{8836465}
Dai, J., Chen, C. \& Li, Y. (2019).
\newblock A Backdoor Attack Against LSTM-Based Text Classification Systems. 
\newblock {\em IEEE Access}. 

\bibitem[{Liu, K. et~al.(2018) Liu, K., Dolan-Gavitt, B. \& Garg, S.}]{Fine-Pruning}
Liu, K., Dolan-Gavitt, B. \& Garg, S. (2018).
\newblock Fine-Pruning: Defending Against Backdooring Attacks on Deep Neural Networks. 
\newblock {\em RAID}. 

\bibitem[{Guo, W. et~al.(2019) Guo, W., Wang, L., Xing, X., Du, M. \& Song, D.}]{TABOR}
Guo, W., Wang, L., Xing, X., Du, M. \& Song, D. (2019)
\newblock TABOR: A Highly Accurate Approach to Inspecting and Restoring Trojan Backdoors in AI Systems. 
\newblock {\em ArXiv}. 1908.01763

\bibitem[{Xiang, Z. et~al.(2020) Xiang, Z., Miller, D. \& Kesidis, G.}]{XiangMK20}
Xiang, Z., Miller, D. \& Kesidis, G. (2020).
\newblock Revealing Backdoors, Post-Training, in DNN Classifiers via Novel Inference on Optimized Perturbations Inducing Group Misclassification. 
\newblock {\em ICASSP}. 

\bibitem[{Shen, Y. et~al.(2019) Shen, Y. \& Sanghavi, S.}]{trim_loss}
Shen, Y. \& Sanghavi, S. (2019).
\newblock Learning with Bad Training Data via Iterative Trimmed Loss Minimization. 
\newblock {\em ICML}. pp. 5739-5748 

\bibitem[{Madry, A. et~al.(2018) Madry, A., Makelov, A., Schmidt, L., Tsipras, D. \& Vladu, A.}]{PGD}
Madry, A., Makelov, A., Schmidt, L., Tsipras, D. \& Vladu, A. (2018).
\newblock Towards Deep Learning Models Resistant to Adversarial Attacks. 
\newblock {\em ICLR}. 

\bibitem[{Doan, B. et~al.(2020) Doan, B., Abbasnejad, E. \& Ranasinghe, D.}]{Februus}
Doan, B., Abbasnejad, E. \& Ranasinghe, D. (2020).
\newblock Februus: Input Purification Defense Against Trojan Attacks on Deep Neural Network Systems. 
\newblock {\em Annual Computer Security Applications Conference}. pp. 897-912 

\bibitem[{Kolouri, S. et~al.(2020) Kolouri, S., Saha, A., Pirsiavash, H. \& Hoffmann, H.}]{meta_sup}
Kolouri, S., Saha, A., Pirsiavash, H. \& Hoffmann, H. (2020).
\newblock Universal Litmus Patterns: Revealing Backdoor Attacks in CNNs. 
\newblock {\em CVPR}. pp. 298-307 

\bibitem[{Xu, X. et~al.(2021) Xu, X., Wang, Q., Li, H., Borisov, N., Gunter, C. \& Li, B.}]{meta_unsup}
Xu, X., Wang, Q., Li, H., Borisov, N., Gunter, C. \& Li, B. (2021).
\newblock Detecting AI Trojans Using Meta Neural Analysis. 
\newblock {\em Proc. IEEE Symposium On Security And Privacy}. 

\bibitem[{Wang, R. et~al.(2020) Wang, R., Zhang, G., Liu, S., Chen, P., Xiong, J. \& Wang, M.}]{DataLimited}
Wang, R., Zhang, G., Liu, S., Chen, P., Xiong, J. \& Wang, M. (2020).
\newblock Practical Detection of Trojan Neural Networks: Data-Limited and Data-Free Cases. 
\newblock {\em ECCV}. 

\bibitem[{Du, M. et~al.(2020) Du, M., Jia, R. \& Song, D.}]{Differential_Privacy}
Du, M., Jia, R. \& Song, D. (2020).
\newblock Robust Anomaly Detection and Backdoor Attack Detection Via Differential Privacy. 
\newblock {\em ICLR}. 

\bibitem[{Li, Y. et~al.(2021) Li, Y., Lyu, X., Koren, N., Lyu, L., Li, B. \& Ma, X.}]{li2021anti}
Li, Y., Lyu, X., Koren, N., Lyu, L., Li, B. \& Ma, X. (2021).
\newblock Anti-Backdoor Learning: Training Clean Models on Poisoned Data. 
\newblock {\em NeurIPS}. 

\bibitem[{Xiang, Z. et~al.(2019) Xiang, Z., Miller, D. \& Kesidis, G.}]{MLSP}
Xiang, Z., Miller, D. \& Kesidis, G. (2019).
\newblock A Benchmark Study Of Backdoor Data Poisoning Defenses For Deep Neural Network Classifiers And A Novel Defense. 
\newblock {\em MLSP}. 

\bibitem[{Liu, Y. et~al.(2019) Liu, Y., Lee, W., Tao, G., Ma, S., Aafer, Y. \& Zhang, X.}]{ABS}
Liu, Y., Lee, W., Tao, G., Ma, S., Aafer, Y. \& Zhang, X. (2019).
\newblock ABS: Scanning Neural Networks for Back-Doors by Artificial Brain Stimulation. 
\newblock {\em Proceedings Of The 2019 ACM SIGSAC Conference On Computer And Communications Security}. pp. 1265-1282 

\bibitem[{Chen, H. et~al.(2019) Chen, H., Fu, C., Zhao, J. \& Koushanfar, F.}]{DeepInspect}
Chen, H., Fu, C., Zhao, J. \& Koushanfar, F. (2019).
\newblock DeepInspect: A Black-box Trojan Detection and Mitigation Framework for Deep Neural Networks. 
\newblock {\em IJCAI}. pp. 4658-4664 

\bibitem[{Dong, Y. et~al.(2021) Dong, Y., Yang, X., Deng, Z., Pang, T., Xiao, Z., Su, H. \& Zhu, J.}]{NC_blackbox}
Dong, Y., Yang, X., Deng, Z., Pang, T., Xiao, Z., Su, H. \& Zhu, J. (2021).
\newblock Black-box Detection of Backdoor Attacks with Limited Information and Data.
\newblock {\em ICCV}.  

\bibitem[{Miller, D. et~al.(2020) Miller, D., Xiang, Z. \& Kesidis, G.}]{Proceedings}
Miller, D., Xiang, Z. \& Kesidis, G. (2020).
\newblock Adversarial Learning Targeting Deep Neural Network Classification: A Comprehensive Review of Defenses Against Attacks. 
\newblock {\em Proc. IEEE}. 

\bibitem[{Tran, B. et~al.(2018) Tran, B., Li, J. \& Madry, A.}]{SS}
Tran, B., Li, J. \& Madry, A. (2018).
\newblock Spectral Signatures in Backdoor Attacks. 
\newblock {\em NeurIPS}. 

\bibitem[{Chen, B. et~al.(2019) Chen, B., Carvalho, W., Baracaldo, N., Ludwig, H., Edwards, B., Lee, T., Molloy, I. \& Srivastava, B.}]{AC}
Chen, B., Carvalho, W., Baracaldo, N., Ludwig, H., Edwards, B., Lee, T., Molloy, I. \& Srivastava, B. (2019).
\newblock Detecting Backdoor Attacks on Deep Neural Networks by Activation Clustering. 
\newblock {\em AAAI}. 

\bibitem[{Simonyan, K. et~al.(2015) Simonyan, K. \& Zisserman, A.}]{VGG}
Simonyan, K. \& Zisserman, A. (2015).
\newblock Very Deep Convolutional Networks for Large-Scale Image Recognition. 
\newblock {\em ICLR}. 

\bibitem[{Lecun, Y. et~al.(1998) Lecun, Y., Bottou, L., Bengio, Y. \& Haffner, P.}]{LeNet}
Lecun, Y., Bottou, L., Bengio, Y. \& Haffner, P. (1998).
\newblock Gradient-based learning applied to document recognition. 
\newblock {\em Proceedings Of The IEEE}. 

\bibitem[{Kumar, N. et~al.(2009) Kumar, N., Berg, A., Belhumeur, P. \& Nayar, S.}]{PubFig}
Kumar, N., Berg, A., Belhumeur, P. \& Nayar, S. (2009).
\newblock Attribute and Simile Classifiers for Face Verification. 
\newblock {\em ICCV}. 

\bibitem[{Deng, L.(2012)}]{MNIST}
Deng, L. (2012).
\newblock The mnist database of handwritten digit images for machine learning research. 
\newblock {\em IEEE Signal Processing Magazine}. 

\bibitem[{Xiao, H. et~al.(2017) Xiao, H., Rasul, K. \& Vollgraf, R.}]{FMNIST}
Xiao, H., Rasul, K. \& Vollgraf, R. (2017).
\newblock Fashion-MNIST: a Novel Image Dataset for Benchmarking Machine Learning Algorithms.  
\newblock {\em ArXiv}. 1708.07747

\bibitem[{Xiang, Z. et~al.(2021) Xiang, Z., Miller, D., Chen, S., Li, X. \& Kesidis, G.}]{ZhenICCV}
Xiang, Z., Miller, D., Chen, S., Li, X. \& Kesidis, G. (2021).
\newblock A Backdoor Attack against 3D Point Cloud Classifiers. 
\newblock {\em ICCV}. 

\bibitem[{Xiang, Z. et~al.(2022) Xiang, Z., Miller, D., Chen, S., Li, X. \& Kesidis, G.}]{XiangMCLK22}
Xiang, Z., Miller, D., Chen, S., Li, X. \& Kesidis, G. (2022).
\newblock Detecting Backdoor Attacks against Point Cloud Classifiers. 
\newblock {\em ICASSP}. 

\bibitem[{Li, Y. et~al.(2017) Li, Y., Wang, N., Shi, J., Liu, J. \& Hou, X.}]{AdaBN}
Li, Y., Wang, N., Shi, J., Liu, J. \& Hou, X. (2017).
\newblock Revisiting Batch Normalization For Practical Domain Adaptation. 
\newblock {\em ICLR}. 

\bibitem[{Ioffe, S. et~al.(2015) Ioffe, S. \& Szegedy, C.}]{BN}
Ioffe, S. \& Szegedy, C. (2015).
\newblock Batch Normalization: Accelerating Deep Network Training by Reducing Internal Covariate Shift. 
\newblock {\em ICML}. 

\bibitem[{Zeng, Y. et~al.(2022) Zeng, Y., Chen, S., Park, W., Mao, Z., Jin, M. \& Jia, R.}]{hypergrad}
Zeng, Y., Chen, S., Park, W., Mao, Z., Jin, M. \& Jia, R. (2022).
\newblock Adversarial Unlearning of Backdoors via Implicit Hypergradient. 
\newblock {\em ICLR}. 

\bibitem[{Ali, S. et~al.(1966) Ali, S. \& Silvey, S.}]{Ali1966AGC}
Ali, S. \& Silvey, S. (1966).
\newblock A General Class of Coefficients of Divergence of One Distribution from Another. 
\newblock {\em Journal Of The Royal Statistical Society Series B-methodological}. \textbf{28} pp. 131-142 

\bibitem[{Li, Y. et~al.(2021) Li, Y., Lyu, X., Koren, N., Lyu, L., Li, B. \& Ma, X.}]{NAD}
Li, Y., Lyu, X., Koren, N., Lyu, L., Li, B. \& Ma, X. (2021).
\newblock Neural Attention Distillation: Erasing Backdoor Triggers from Deep Neural Networks. 
\newblock {\em ICLR}. 

\bibitem[Zhao, P. et~al.(2020) Zhao, P., Chen, P., Das, P., Ramamurthy, K. \& Lin, X.]{MCR}
Zhao, P., Chen, P., Das, P., Ramamurthy, K. \& Lin, X. (2020).
\newblock Bridging Mode Connectivity in Loss Landscapes and Adversarial Robustness. 
\newblock {\em ICLR}. 

\bibitem[{Nguyen, T. et~al.(2021) Nguyen, T. \& Tran, A.}]{WaNet}
Nguyen, T. \& Tran, A. (2021).
\newblock WaNet - Imperceptible Warping-based Backdoor Attack. 
\newblock {\em ICLR}. 

\bibitem[{Li, S. et~al.(2021) Li, S., Xue, M., Zhao, B., Zhu, H. \& Zhang, X.}]{invisible}
Li, S., Xue, M., Zhao, B., Zhu, H. \& Zhang, X. (2021).
\newblock Invisible Backdoor Attacks on Deep Neural Networks Via Steganography and Regularization. 
\newblock {\em IEEE Transactions On Dependable And Secure Computing}. \textbf{18}, 2088-2105 

\bibitem[{Huang, K. et~al.(2022) Huang, K., Li, Y., Wu, B., Qin, Z. \& Ren, K.}]{huang2022backdoor}
Huang, K., Li, Y., Wu, B., Qin, Z. \& Ren, K. (2022).
\newblock Backdoor Defense via Decoupling the Training Process. 
\newblock {\em ICLR}. 

\bibitem[{Wu, D. et~al.(2021) Wu, D. \& Wang, Y.}]{ANP}
Wu, D. \& Wang, Y. (2021).
\newblock Adversarial Neuron Pruning Purifies Backdoored Deep Models. 
\newblock {\em NeurIPS}. 

\bibitem[{Zheng, R. et~al.(2022) Zheng, R., Tang, R., Li, J. \& Liu, L.}]{CLP}
Zheng, R., Tang, R., Li, J. \& Liu, L. (2022).
\newblock Data-free Backdoor Removal based on Channel Lipschitzness. 
\newblock {\em ECCV}. 

\bibitem[{Guan, J. et~al.(2022) Guan, J., Tu, Z., He, R. \& Tao, D.}]{ShapPruning}
Guan, J., Tu, Z., He, R. \& Tao, D. (2022).
\newblock Few-shot Backdoor Defense Using Shapley Estimation. 
\newblock {\em CVPR}. 

\bibitem[{Xia, J. et~al.(2022) Xia, J., Wang, T., Ding, J., Wei, X. \& Chen, M.}]{ARGD}
Xia, J., Wang, T., Ding, J., Wei, X. \& Chen, M. (2022).
\newblock Eliminating Backdoor Triggers for Deep Neural Networks Using Attention Relation Graph Distillation. 
\newblock {\em IJCAI}. 

\bibitem[{Wang, H. et~al.(2022) Wang, H., Xiang, Z., Miller, D. \& Kesidis, G.}]{UniBD}
Wang, H., Xiang, Z., Miller, D. \& Kesidis, G. (2022).
\newblock Universal Post-Training Backdoor Detection. 
\newblock {\em ArXiv}. \textbf{2205.069} 

\bibitem[{Bai, J. et~al.(2022) Bai, J., Gao, K., Gong, D., Xia, S., Li, Z. \& Liu, W.}]{Bai_2022_ECCV}
Bai, J., Gao, K., Gong, D., Xia, S., Li, Z. \& Liu, W. (2022).
\newblock Hardly Perceptible Trojan Attack against Neural Networks with Bit Flips. 
\newblock {\em ECCV}. 

\bibitem[{Liu, Y. et~al.(2022) Liu, Y., Shen, G., Tao, G., Wang, Z., Ma, S. \& Zhang, X.}]{Liu_2022_CVPR}
Liu, Y., Shen, G., Tao, G., Wang, Z., Ma, S. \& Zhang, X. (2022).
\newblock Complex Backdoor Detection by Symmetric Feature Differencing. 
\newblock {\em CVPR}. 

\bibitem[{Barni, M. et~al.(2019) Barni, M., Kallas, K. \& Tondi, B.}]{SIG}
Barni, M., Kallas, K. \& Tondi, B. (2019).
\newblock A New Backdoor Attack in {CNNS} by Training Set Corruption Without Label Poisoning.
\newblock {\em ICIP}

\bibitem[{Tao, G. et~al.(2022) Tao, G., Shen, G., Liu, Y., An, S., Xu, Q., Ma, S., Li, P. \& Zhang, X.}]{Tao_2022_CVPR}
Tao, G., Shen, G., Liu, Y., An, S., Xu, Q., Ma, S., Li, P. \& Zhang, X. (2022).
\newblock Better Trigger Inversion Optimization in Backdoor Scanning. 
\newblock {\em CVPR}. 

\bibitem[{Houben, S. et~al.(2013) Houben, S., Stallkamp, J., Salmen, J., Schlipsing, M. \& Igel, C.}]{GTSRB}
Houben, S., Stallkamp, J., Salmen, J., Schlipsing, M. \& Igel, C. (2013).
\newblock Detection of Traffic Signs in Real-World Images: The German Traffic Sign Detection Benchmark. 
\newblock {\em International Joint Conference On Neural Networks}. 

\bibitem[{Howard, J. (2020)}]{imagewang}
Howard, J. (2020). 
\newblock ImageNette. 
\newblock \url{https://github.com/fastai/imagenette/}

\bibitem[{Le, Y. et~al.(2015) Le, Y. \& Yang, X.}]{TinyImageNet}
Le, Y. \& Yang, X. (2015).
\newblock Tiny ImageNet Visual Recognition Challenge. 
\newblock \url{https://tiny-imagenet.herokuapp}.

\bibitem[Cao, Q. et~al.(2018) Cao, Q., Shen, L., Xie, W., Parkhi, O. \& Zisserman, A.]{vggface2}
Cao, Q., Shen, L., Xie, W., Parkhi, O. \& Zisserman, A. (2018).
\newblock VGGFace2: A dataset for recognising faces across pose and age. 
\newblock {\em International Conference On Automatic Face And Gesture Recognition}. 

\bibitem[{Russakovsky, O. et~al.(2015) Russakovsky, O., Deng, J., Su, H., Krause, J., Satheesh, S., Ma, S., Huang, Z., Karpathy, A., Khosla, A., Bernstein, M., Berg, A. \& Fei-Fei, L.}]{ImageNet}
Russakovsky, O., Deng, J., Su, H., Krause, J., Satheesh, S., Ma, S., Huang, Z., Karpathy, A., Khosla, A., Bernstein, M., Berg, A. \& Fei-Fei, L. (2015).
\newblock ImageNet Large Scale Visual Recognition Challenge. 
\newblock {\em Int. J. Comput. Vis.}. \textbf{115}, 211-252 



\end{thebibliography}

\end{document}